# A New General Method to Generate Random Modal Formulae for Testing Decision Procedures

**Peter F. Patel-Schneider**                                              PFPS@RESEARCH.BELL-LABS.COM
*Bell Labs Research*
*600 Mountain Ave. Murray Hill, NJ 07974, USA*

**Roberto Sebastiani**                                                         RSEBA@DIT.UNITN.IT
*Dip. di Informatica e Telecomunicazioni*
*Università di Trento*
*via Sommarive 14, I-38050, Trento, Italy*

## Abstract

The recent emergence of heavily-optimized modal decision procedures has highlighted the key role of empirical testing in this domain. Unfortunately, the introduction of extensive empirical tests for modal logics is recent, and so far none of the proposed test generators is very satisfactory. To cope with this fact, we present a new random generation method that provides benefits over previous methods for generating empirical tests. It fixes and much generalizes one of the best-known methods, the random $CNF_{\Box_m}$ test, allowing for generating a much wider variety of problems, covering in principle the whole input space. Our new method produces much more suitable test sets for the current generation of modal decision procedures. We analyze the features of the new method by means of an extensive collection of empirical tests.

## 1. Motivation and Goals

Heavily-optimized systems for determining satisfiability of formulae in propositional modal logics are now available. These systems, including DLP (Patel-Schneider, 1998), FACT (Horrocks, 1998), *SAT (Giunchiglia, Giunchiglia, & Tacchella, 2002), MSPASS (Hustadt, Schmidt, & Weidenbach, 1999), and RACER (Haarslev & Möller, 2001), have more optimizations and are much faster than the previous generation of modal decision procedures, such as LEANK (Beckert & Goré, 1997), LOGICS WORKBENCH (Heuerding, Jäger, Schwendimann, & Seyfreid, 1995), □KE (Pitt & Cunningham, 1996) and KSAT (Giunchiglia & Sebastiani, 2000).[1]

As with most theorem proving problems, neither computational complexity nor asymptotic algorithmic complexity is very useful in determining the effectiveness of optimizations, so that their effectiveness has to be determined by empirical testing (Horrocks, Patel-Schneider, & Sebastiani, 2000). Empirical testing directly gives resource consumption in terms of compute time and memory use; it factors in all the pieces of the system, not just the basic algorithm itself. Empirical testing can be used not only to compare different systems, but also to tune a system with parameters that can be used to modify its performance; moreover, it can be used to show what sort of inputs the system handles well, and what sort of inputs the system handles poorly.

Unfortunately, the introduction of extensive empirical tests for modal logics is recent, and so far none of the proposed test methodologies are very satisfactory. Some methods contain many formulae that are too easy for current heavily-optimized procedures. Some contain high rates of trivial or insignificant tests. Some generate problems that are too artificial and/or are not a significant sample of the input space. Finally, some methods generate formulae that are too big to be parsed and/or handled.

---

1. For a more complete list see Renate Schmidt's Web page listing theorem provers for modal logics at http://www.cs.man.ac.uk/~schmidt/tools/.

For the reasons described above, we presented (Horrocks et al., 2000) an analytical survey of the state-of-the art of empirical testing for modal decision procedures. Here instead we present a new random generation method that provides benefits over previous methods for generating empirical tests, built on some preliminary work (Horrocks et al., 2000). Our new method fixes and much generalizes the $3CNF_{\square_m}$ methodology for randomly generating clausal formulae in modal logics (Giunchiglia & Sebastiani, 1996; Hustadt & Schmidt, 1999; Giunchiglia, Giunchiglia, Sebastiani, & Tacchella, 2000) used in many previous empirical tests of modal decision procedures. It eliminates or drastically reduces the influence of a major flaw of the previous method,[2] and allows for generating a much wider variety of problems.

In Section 2 we recall a list of desirable features for good test sets. In Section 3 we briefly survey the state-of-the-art test methods. In Sections 4 and 5 we present and discuss the basic and the advanced versions of our new test method respectively, and evaluate their features by presenting a large amount of empirical results. In Section 6 we provide a theoretical result showing how the advanced version of our method, in principle, can cover the whole input space. In Section 7 we discuss the features of our new method, and compare it wrt. the state-of-the-art methods. In Section 8 we conclude and indicate possible future research directions.

A 5-page system description of our random generator has been presented at IJCAR'2001 (Patel-Schneider & Sebastiani, 2001).

## 2. Desirable Features for Good Test Sets

The benefits of empirical testing depend on the characteristics of the inputs provided for the testing, as empirical testing only provides data on these particular inputs. If the inputs are not typical or suitable, then the results of the empirical testing will not be useful. This means that the inputs for empirical testing must be carefully chosen. With Horrocks (Horrocks et al., 2000) we have previously proposed and motivated the following key criteria for creating good test sets.

**Representativeness:** The ideal test set should represent a significant sample of the whole input space. A good empirical test set should at least cover a large area of inputs.

**Difficulty:** A good empirical test set should provide a sufficient level of difficulty for the system(s) being tested. (Some problems should be too hard even for state-of-the-art systems, so as to be a good benchmark for forthcoming systems.)

**Termination:** To be of practical use, the tests should terminate and provide information within a reasonable amount of time. If the inputs are too hard, then the system may not be able to provide answers within the established time. This inability of the system is of interest, but can make system comparison impossible or insignificant.

---

2. That is, a significant amount of inadvertently trivial problems are generated unless the parameter p is set to 0 (Horrocks et al., 2000). See Section 4.1 for a full discussion of this point.

**Scalability:** The difficulty of problems should scale up, as comparing absolute performances may be less significant than comparing how performances scale up with problems of increasing difficulty.

**Valid vs. not-valid balance:** In a good test set, valid and not-valid problems should be more or less equal both in number and in difficulty. Moreover, the *maximum uncertainty* regarding the solution of the problems is desirable.

**Reproducibility:** A good test set should allow for easily reproducing the results.

The following criteria derive from or are significant sub-cases of the main criteria above.

**Parameterization:** Parameterized inputs with sufficient parameters and degrees of freedom allow the inputs to range over a large portion of the input space.

**Control:** In particular, it is very useful to have parameters that control *monotonically* the key features of the input test set, like the average difficulty and the "valid vs. non-valid" rate.

**Modal vs. propositional balance:** Reasoning in modal logics involves alternating between two orthogonal search efforts: pure modal reasoning and pure propositional reasoning. A good test set should be challenging from both viewpoints.

**Data organization:** The data should be summarizable —so as to make a comparison possible with a limited effort— and plottable —so as to enable the qualitative behavior of the system(s) to be highlighted.

Finally, particular care must be taken to avoid the following problems.

**Redundancy:** Empirical test sets must be carefully chosen so as not to include inadvertent redundancy. They should also be chosen so as not to include small sub-inputs that dictate the result of the entire input.

**Triviality:** A good test set should be flawless, that is, it should not contain  significant subsets of inadvertent trivial problems.

**Artificiality:** A good empirical test set should correspond closely to inputs from applications.

**Over-size:** The single problems should not be too big w.r.t. their difficulty, so that the resources required for parsing and data managing do not seriously influence total performance.

These criteria, which are described and motivated in detail by Horrocks et al. (2000), have been proposed after a five-year debate on empirical testing in modal logics (Giunchiglia & Sebastiani, 1996; Heuerding & Schwendimann, 1996; Hustadt & Schmidt, 1999; Giunchiglia et al., 2000; Horrocks & Patel-Schneider, 2002). (Notice that some of these criteria are identical or similar to those suggested by Heuerding & Schwendimann, 1996.)

The above criteria are general, and in some cases they require some interpretation. First, some of them have to be implicitly interpreted as "unless the user deliberately wants the contrary for some reason". For instance, it might be the case that one wants to *deliberately* generate easy problems, e.g., to be sure that the tested procedure does not take too much time to solve them, or redundant problems, e.g., to test the effectiveness of some redundancy elimination technique, or satisfiable problems only, e.g., to test incomplete procedures. To this extent, the key issue here is having a reasonable form of control over these features, so that one can address not only general-purpose criteria, but also specific desiderata.

Second, in some cases, there may be a tradeoff between two distinct criteria, so that it may be necessary to choose only one of them, or to make a compromise. One example is given by redundancy and artificiality: in some real-world problems large parts of the knowledge base are irrelevant for the query, whose result is determined by a small subpart of the input; in this sense eliminating such "redundancies" may make problems more "artificial".

Particular attention must be paid to the problem of triviality, as it has claimed victims in many areas of AI. In fact, flaws (i.e., inadvertent trivial problems) have been detected in random generators for SAT (Mitchell, Selman, & Levesque, 1992), CSP (Achlioptas, Kirousis, Kranakis, Krizanc, Molloy, & Stamatiou, 1997; Gent, MacIntyre, Prosser, Smith, & Walsh, 2001), modal reasoning (Hustadt & Schmidt, 1999) and QBF (Gent & Walsh, 1999). Thus, the notion of "trivial" (and thus "flawed") deserves more comment.

In the work by Achlioptas et al. (1997) flawed problems are those solvable in linear time by standard CSP procedures, due to the undesired presence of implicit unary constraints causing some variable's value to be inadmissible. A similar notion holds for SAT (Mitchell et al., 1992) and QBF (Gent & Walsh, 1999). In the literature of modal reasoning, instead, the typical flawed problems are those whose (un)satisfiability can be verified directly at propositional level, that is, without investigating any modal successors; this kind of problems are typically solved in negligible time w.r.t. other problems of similar size and depth (Hustadt & Schmidt, 1999; Giunchiglia et al., 2000; Horrocks et al., 2000).[3] Thus, with a little abuse of notation and when not otherwise specified, in this paper we will call *trivially (un)satisfiable* the problems of this kind.[4]

## 3. An Overview of the State-of-the-art

Previous empirical tests have mostly been generated by three methods: hand-generated formulae (Heuerding & Schwendimann, 1996), randomly-generated clausal modal formulae (Giunchiglia & Sebastiani, 1996; Hustadt & Schmidt, 1999; Giunchiglia et al., 2000), and randomly-generated quantified boolean formulae that are then translated into modal formulae (Massacci, 1999).

We have already presented a detailed analysis of these three methods (Horrocks et al., 2000). Here we present only a quick overview of the latter two methods, as we will refer to them in following sections.[5]

### 3.1 The $3CNF_{\Box_m}$ Random Tests

In the $3CNF_{\Box_m}$ test methodology (Giunchiglia & Sebastiani, 1996; Hustadt & Schmidt, 1999; Giunchiglia et al., 2000), the performance of a system is evaluated on sets of randomly generated $3CNF_{\Box_m}$ formulae. A $CNF_{\Box_m}$ formula is a conjunction of $CNF_{\Box_m}$ clauses, where each clause is a disjunction of either propositional or modal literals. A literal is either an atom or its negation. Modal atoms are formulae of the form $\Box_r C$, where $C$ is a $CNF_{\Box_m}$ clause. A $3CNF_{\Box_m}$ formula is a $CNF_{\Box_m}$ formula where all clauses have exactly 3 literals.

---

3. Of course here by "modal" we implicitly assume the modal depth be strictly greater than zero, that is, we do not consider purely propositional formulas.

4. Notice that we do not use the more suitable expression "propositionally (un)satisfiable" because the latter has been used with a different meaning in the literature of modal reasoning (see, e.g., Giunchiglia & Sebastiani, 1996, 2000).

5. The first method (Heuerding & Schwendimann, 1996) is obsolete, as the formulae generated are too easy for current state-of-the-art deciders (Horrocks et al., 2000).

### 3.1.1 The Random Generator

A $3CNF_{\Box_m}$ formula is randomly generated according to five parameters: the (maximum) modal depth $d$; the number of clauses in the top-level conjunction $L$; the number of propositional variables $N$; the number of distinct box symbols $m$; and the probability $p$ of an atom occurring in a clause at depth $< d$ being purely propositional.

The random $3CNF_{\Box_m}$ generator, in its final version (Giunchiglia et al., 2000), works as follows:

- a $3CNF_{\Box_m}$ formula of depth $d$ is produced by randomly generating $L$ $3CNF_{\Box_m}$ clauses of depth $d$, and forming their conjunction;

- a $3CNF_{\Box_m}$ clause of depth $d$ is produced by randomly generating three distinct, under commutativity of disjunction, $3CNF_{\Box_m}$ atoms of depth $d$, negating each of them with probability 0.5, and forming their disjunction;

- a propositional atom is produced by picking randomly an element of $\{A_1, \ldots, A_N\}$ with uniform probability;

- a $3CNF_{\Box_m}$ atom of depth $d > 0$ is produced by generating with probability $p$ a random propositional atom, and with probability $1 - p$ a $3CNF_{\Box_m}$ atom $\Box_r C$, where $\Box_r$ is picked randomly in $\{\Box_1, \ldots, \Box_m\}$ and $C$ is a randomly generated $3CNF_{\Box_m}$ clause of depth $d - 1$.

Recently Horrocks and Patel-Schneider (2002) have proposed a variant of the $3CNF_{\Box_m}$ random generator of Giunchiglia et al. (2000). They added four extra parameters: $n_p$ and $n_m$, representing respectively the probability that a propositional and modal atom is negated, and $c_{min}$ and $c_{max}$, representing respectively the minimum and maximum number of modal literals in a clause, with equal probability for each number in the range. For their experiments, they always set $n_p = 0.5$ and $c_{min} = c_{max} = 3$. To this extent, $3CNF_{\Box_m}$ formulas can be generated as in the generator of Giunchiglia et al. (2000) by setting $n_p = n_m = 0.5$ and $c_{min} = c_{max} = 3$.

### 3.1.2 Test Method & Data Analysis

The $3CNF_{\Box_m}$ test method works as follows. A typical problem set is characterized by a fixed $N$, $m$, $d$ and $p$: $L$ is varied in such a way as to empirically cover the "100% satisfiable—100% unsatisfiable" transition. Then, for each tuple of the parameters' values (*data point* from now on) in a problem set, a certain number of $3CNF_{\Box_m}$ formulae are randomly generated, and the resulting formulae are given in input to the procedure under test, with a maximum time bound. Satisfiability rates, median/percentile values of the CPU times, and median/percentile values of other parameters, e.g., number of steps, memory, etc., are plotted against the number of clauses $L$ or the ratio of clauses to propositional variables $L/N$.

## 3.2 The Random QBF Tests

In QBF-based benchmarks (such as part of the TANCS'99 benchmarks (Massacci, 1999)), system performances are evaluated on sets of random quantified boolean formulae, which are gener-ated according to the method described by Cadoli, Giovanardi, and Schaerf (1998) and Gent and Walsh (1999) and then converted into modal logic by using a variant of the conversion by Halpern and Moses (1992).

### 3.2.1 THE RANDOM GENERATOR

Random QBF formulae are generated with alternation depth $D$ and at most $V$ variables at each alternation. The matrix is a random propositional CNF formula with $C$ clauses of length $K$, with some constraints on the number of universally and existentially quantified variables within each clause. (This avoids the problem of generating flawed random QBF formulae highlighted by Gent & Walsh, 1999.) For instance, a random QBF formula with $D = 3$, $V = 2$ looks like:

$$\forall v_{32} v_{31}.\exists v_{22} v_{21}.\forall v_{12} v_{11}.\exists v_{02} v_{01}.\psi[v_{32}, ..., v_{01}]. \tag{1}$$

Here $\psi$ is a random CNF formula with parameters $C$, $V$ and $D$. We will denote with $U$ and $E$ the total number of universally and existentially quantified variables respectively. Clearly, both $U$ and $E$ are $O(D \cdot V)$. Moreover, $\varphi$ is the modal formula resulting from Halpern and Moses' **K** conversion, so both the depth $d$ and the number of propositional variables $N$ of $\varphi$ are also $O(D \cdot V)$.

### 3.2.2 TEST METHOD & DATA ANALYSIS

The test method, as it was used in the TANCS competition(s) (Massacci, 1999), works as follows. The tests are performed on single data points. For each data point, a certain number of QBF formulae are randomly generated, converted into modal logics and the resulting formulae are given as input to the procedure being tested, with a maximum time bound. The number of tests which have been solved within the time-limit and the geometrical mean time for successful solutions are then reported. Data are rescaled to abstract away machine and run-dependent characteristics. This results typically in a collection of tables presenting a data pair for each system under test, one data point per row.

## 4. A New CNF$_{\Box_m}$ Generation Method: Basic Version

From our previous analysis (Horrocks et al., 2000) we have that none of the current methods are completely satisfactory. To cope with this fact, we propose here what we believe is a much more satisfactory method for randomly generating modal formulae. The new method can be seen as an improved and much more general version of the random 3CNF$_{\Box_m}$ generation method by Giunchiglia et al. (2000).

We present our new method by introducing incrementally its new features in two main steps. In this section we introduce a *basic version* of the method, wherein

- we provide a new interpretation for the parameter $p$ (Section 4.1) that allows for varying $p$ without causing the flaws described in Horrocks et al. (2000); and

- we extend the interpretation for the parameter $C$ (Section 4.3), providing a more fine-grained way for tuning the difficulty of the generated formulae.

In Section 5, we present the full, *advanced version* of the method, wherein

- we further extend the parameters $p$ and $C$, allowing for shaping explicitly the probability distribution of the propositional/modal rate and the clause length respectively (Section 5.1); and

- we allow $p$ and $C$ vary with the nesting depth of the subformulae (Section 5.2), allowing for different distributions at different depths.

To investigate the properties of our $\text{CNF}_{\square_m}$ generator we also present a series of experiments with appropriate settings either to mimic previous generation methodologies or to produce improved or new kinds of tests.

In all tests we have adopted the testing criteria of the $3\text{CNF}_{\square_m}$ method. For each test set, we fixed all parameters except $L$, which was varied to span at least the satisfiability transition area. (Because of the "Valid vs. non-valid balance" feature of Section 2, we consider the transition area to be the interesting portion of the test set.) For almost all test sets we varied $L$ from $N$ to $120N$, $150N$, or $200N$, resulting in integral values for $L/N$ ranging from 1 to 120, 150, or 200. For each value of $L$ we generated 100 formulae, a sufficient number to produce reasonably reliable data. A time limit of 1000 seconds was imposed on each attempt to determine the satisfiability status of a formula. As it is common practice, we set the number of boxes $m$ to 1 throughout our testing. This setting for $m$ produces the hardest formulae (Giunchiglia & Sebastiani, 1996; Hustadt & Schmidt, 1999; Giunchiglia et al., 2000). We performed several test sets with similar parameters, often, but not always, varying only $N$.

We tested our formulae against two systems, DLP version 4.1 (Patel-Schneider, 1998) and *SAT version 1.3 (Tacchella, 1999), two of the fastest modal decision procedures. They are available at http://www.bell-labs.com/usr/pfps/dlp and http://www.mrg.dist.unige.it/~tac respectively. All the code used to generate the tests is available at http://www.bell-labs.com/usr/pfps/dlp.

We plotted the results of our test groups (test sets with similar parameters) on six or four plots. Two plots were devoted to the performance of DLP, one showing the median and one showing the 90th percentile time taken to solve the formulae at each value of $L$, plotted against $L/N$. For those test groups were we ran *SAT we also plotted the median and 90th percentile for *SAT.

We also plotted the fraction of the formulae that are determined to be satisfiable or unsatisfiable by DLP within the time limit.[6] To save space, satisfiability and unsatisfiability fractions are plotted together on a single plot. Satisfiability fractions are higher on the left side of the plot while unsatisfiability fractions are higher on the right. This multiple plotting does obscure some of the details, but the only information that we are interested in here is the general behavior of the fractions, which is not obscured. In fact, the multiple plotting serves to highlight the crossover regions, where the satisfiability and unsatisfiability fractions are roughly equal.

Finally, we plotted the fraction of the formulae where DLP finds a model or determines that the formula is unsatisfiable without investigating any modal successors. We call these fractions the trivial satisfiability and trivial unsatisfiability fractions. These last fractions are an estimate of the number of formulae that are satisfiable in a Kripke structure with no successors —like, e.g., $(A_1 \vee \square_1 A_2)$— and that have no propositional valuations —like, e.g., $(\square_1 A_1 \wedge \neg \square_1 A_1)$— respectively. For various reasons, discussed below, they are better indicators of triviality than the more formal measures used in previous papers. Again, trivial satisfiability and unsatisfiability fractions are plotted together on a single plot.

---

6. Notice that the two curves are symmetric with respect to 0.5 if and only if no test exceeds the time limit. E.g., if at some point 40% of the tests are determined to be satisfiable by DLP, 10% are determined to be unsatisfiable and 50% are not solved within the time limit, then the two curves are not symmetric at that point, as $0.40 \neq 1 - 0.10$.

To reduce clutter on the plots, we used a line to show the results for each value of $L$ we tested. To distinguish between the various lines on a plot, we plotted every five or 10 data points with a symbol, identified in the legend of the plot.

Running the tests presented in this paper required some months of CPU time. Because of this, we ran our tests on a variety of machines. These machines range in speed from a 296MHz SPARC Ultra 2 to a 400MHz SPARC Ultra 4 and had between 256MB and 512MB of main memory. No machines were completely dedicated to our tests, but they were otherwise lightly loaded. Each test set was run on machines with the same speed and memory. Direct comparison between different groups of tests thus has to take into account the differences between the various test machines.

## 4.1 Reinterpreting the Parameter $p$

One problem with the previous methods for generating $CNF_{\Box_m}$ formulae is that the generated formulae can contain pieces that make the entire formula easy to solve. This mostly results from the presence of strictly-propositional top-level clauses. With the small number of propositional variables in most tests (required to produce reasonable difficulty levels for current systems), only a few strictly-propositional top-level clauses are needed to cover all the combinations of the propositional literals and make the entire formula unsatisfiable. Previous attempts to eliminate this "trivial unsatisfiability" have concentrated on eliminating top-level propositional literals by setting $p = 0$ (Hustadt & Schmidt, 1999; Giunchiglia et al., 2000). (Unfortunately this choice forces $d \leq 1$, as for $d > 1$ such formulae are too hard for all state-of-the-art systems.) When each atom in a clause is generated independently from the other atoms of the clause an approach that modifies the probability of propositional atoms is necessary to eliminate these problematic clauses.

The first new idea of our approach, suggested previously (Horrocks et al., 2000), works as follows. Instead of forbidding strictly-propositional clauses except at the maximum modal depth, $d$, by setting $p = 0$, we instead require that the ratio between propositional atoms in a clause and the clause size be as close as possible to the propositional probability $p$ for clauses not at the maximum modal depth $d$. [7]

For clauses of size $C$, if $p$ is $k/C$ for some integral $k$, this results in all clauses not at modal depth $d$ having $k$ propositional atoms and $C - k$ modal atoms. For other values of $p$, we allow either $\lfloor pC \rfloor$ or $\lceil pC \rceil$ propositional atoms in each clause not at modal depth $d$, with probability $\lceil pC \rceil - pC$ and $pC - \lfloor pC \rfloor$, respectively. [8] For instance, if $p = 0.6$ and $C = 3$, then each clause contains 1 propositional and 1 modal literal, and the third is propositional with probability 0.8, as $3 \cdot 0.6 - \lfloor 3 \cdot 0.6 \rfloor = 1.8 - 1 = 0.8$. If $p \leq (C - 1)/C$, this eliminates the possibility of strictly propositional clauses, which are the main cause of trivial unsatisfiability, except at modal depth $d$.

---

7. Other approaches to eliminating propositional unsatisfiability are possible. For example, it would be possible to simply remove any strictly-propositional clauses after generation. However, this technique would alter the meaning of the parameter $p$, that is, the actual probability for a literal to be propositional would become strictly smaller than $p$, and it will be out of the control of the user.

8. Remember that $\lfloor x \rfloor =_{def} max\{n \in \mathcal{N} | n \leq X\}$ and $\lceil x \rceil =_{def} min\{n \in \mathcal{N} | n \geq X\}$.

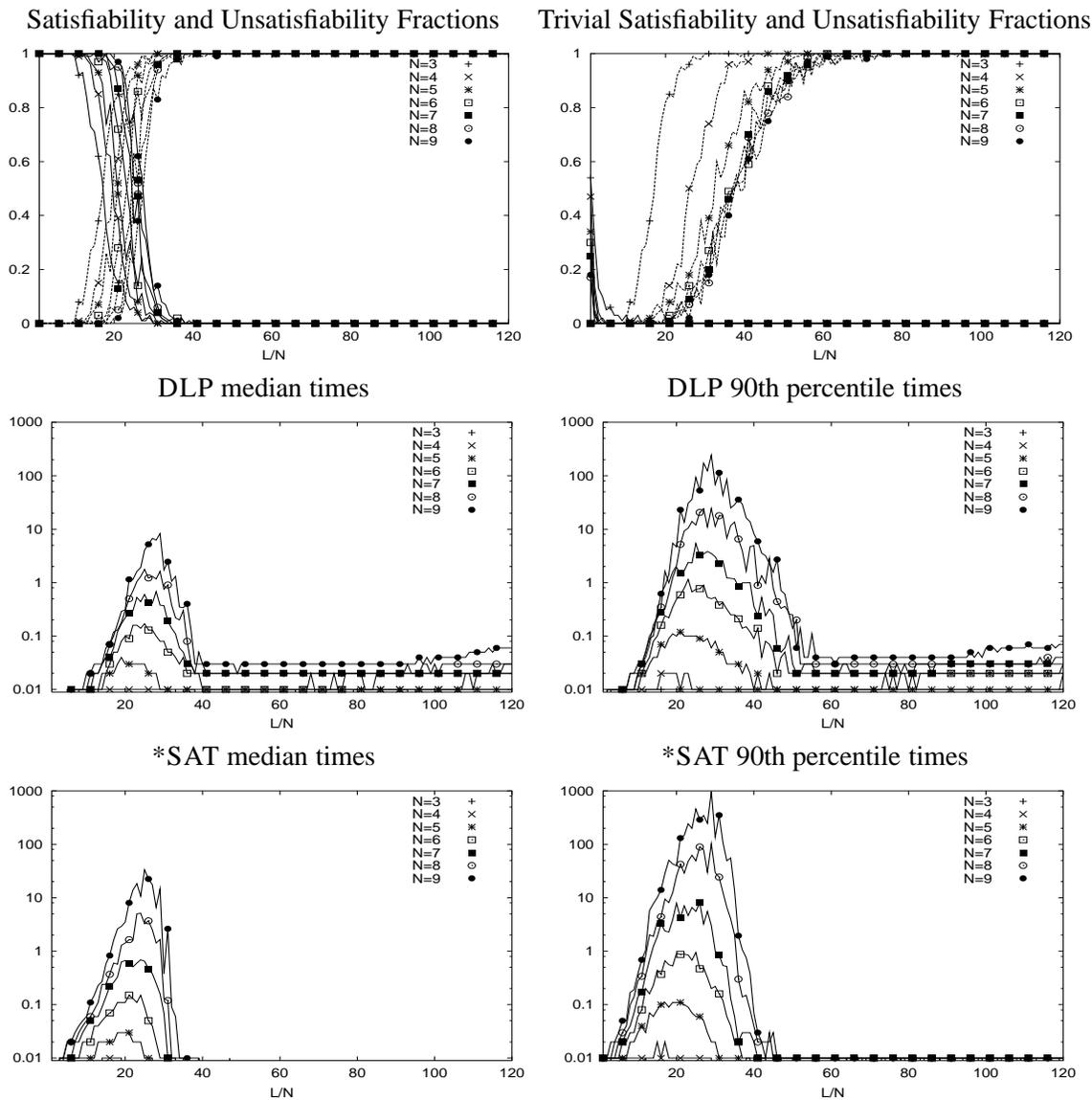

Figure 1: Results for $C = 3$, $m = 1$, $d = 1$, and $p = 0.5$ (old method)

### 4.1.1 Modal Depth 1

Our first experiments were a direct comparison to previous tests. We generated CNF$_{\Box_m}$ formulae with $C = 3$, $m = 1$, $d = 1$, and $p = 0.5$, a setting that has been used in the past, and one that exhibits some problematic behavior. We used both our new method and the old 3CNF$_{\Box_m}$ generation method by Giunchiglia et al. (2000) briefly described in Section 3.1 (the "old method" from now on). We also generated CNF$_{\Box_m}$ formulae with $C = 3$, $m = 1$, $d = 1$, and $p = 0$, the standard method for eliminating trivially unsatisfiable formulae. (At $p = 0$ our new method is the same as the old 3CNF$_{\Box_m}$ generation method.) The results of the tests are given in Figures 1, 2, and 3.

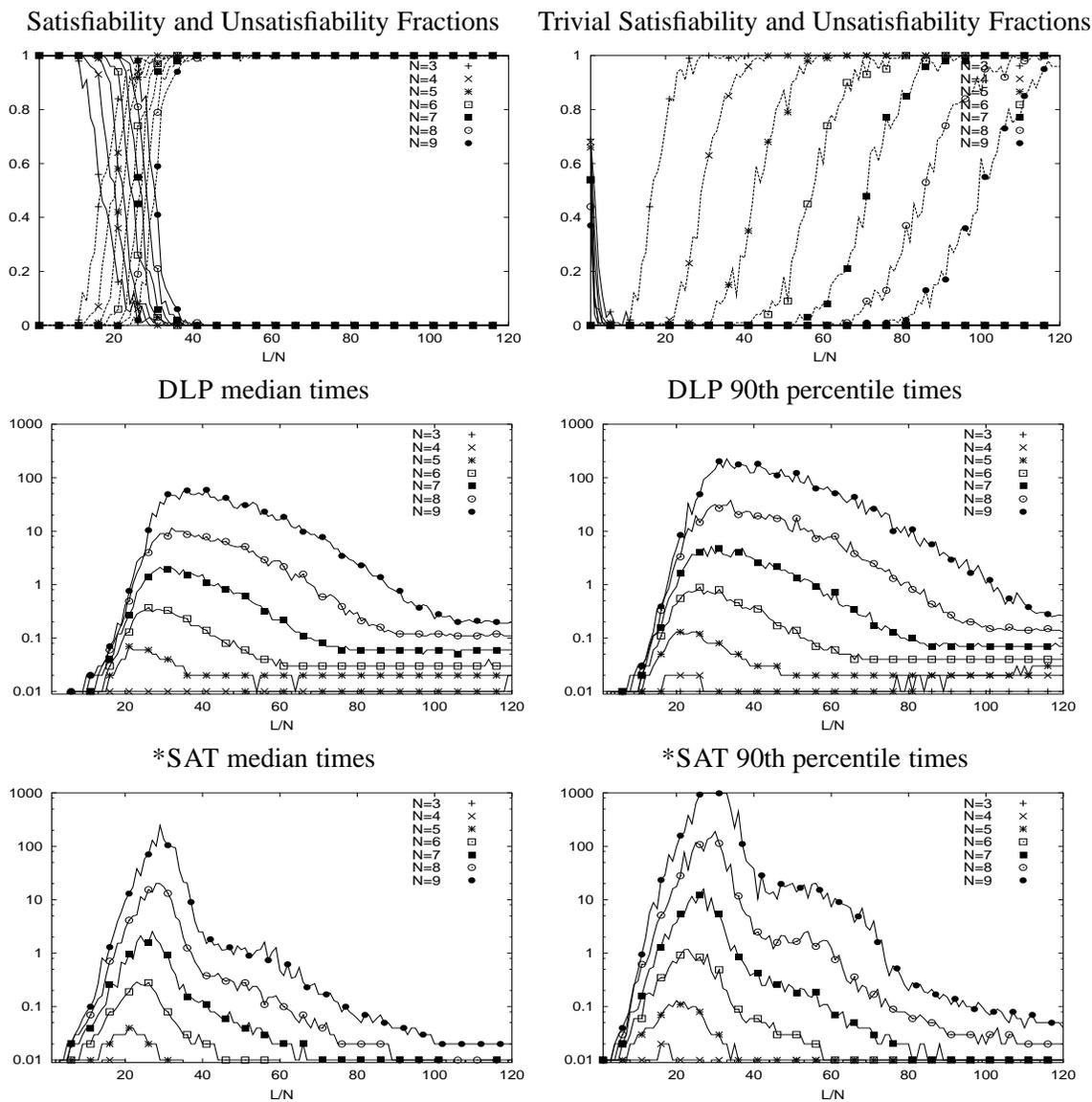

Figure 2: Results for $C = 3$, $m = 1$, $d = 1$, and $p = 0.5$ (our new method)

One aspect of this set of tests is that all three collections have many trivially unsatisfiable formulae out of the satisfiability transition area, even the collection with no top-level propositional atoms. The trivial unsatisfiability occurs in the collection with no top-level propositional atoms because there are only a few top-level modal atoms (e.g., 8 for $N = 3$) and both DLP and *SAT detect clashes between complementary modal literals without investigating any modal successors.

The presence of this large number of trivially unsatisfiable formulae is not actually a serious problem with *these* tests. The trivial unsatisfiability only shows up after the formulae are almost all unsatisfiable already and easy to solve. The only exception is for $N = 3$, which is trivial to solve anyway. However, our new generation method considerably reduces the number of trivially unsatisfiable formulae and almost entirely removes them from the satisfiable/unsatisfiable transition area. There are some trivially satisfiable formulae in this set of tests, but only a few, and only for the smallest clause sizes. Their presence does not affect the difficulty of the generated formulae.

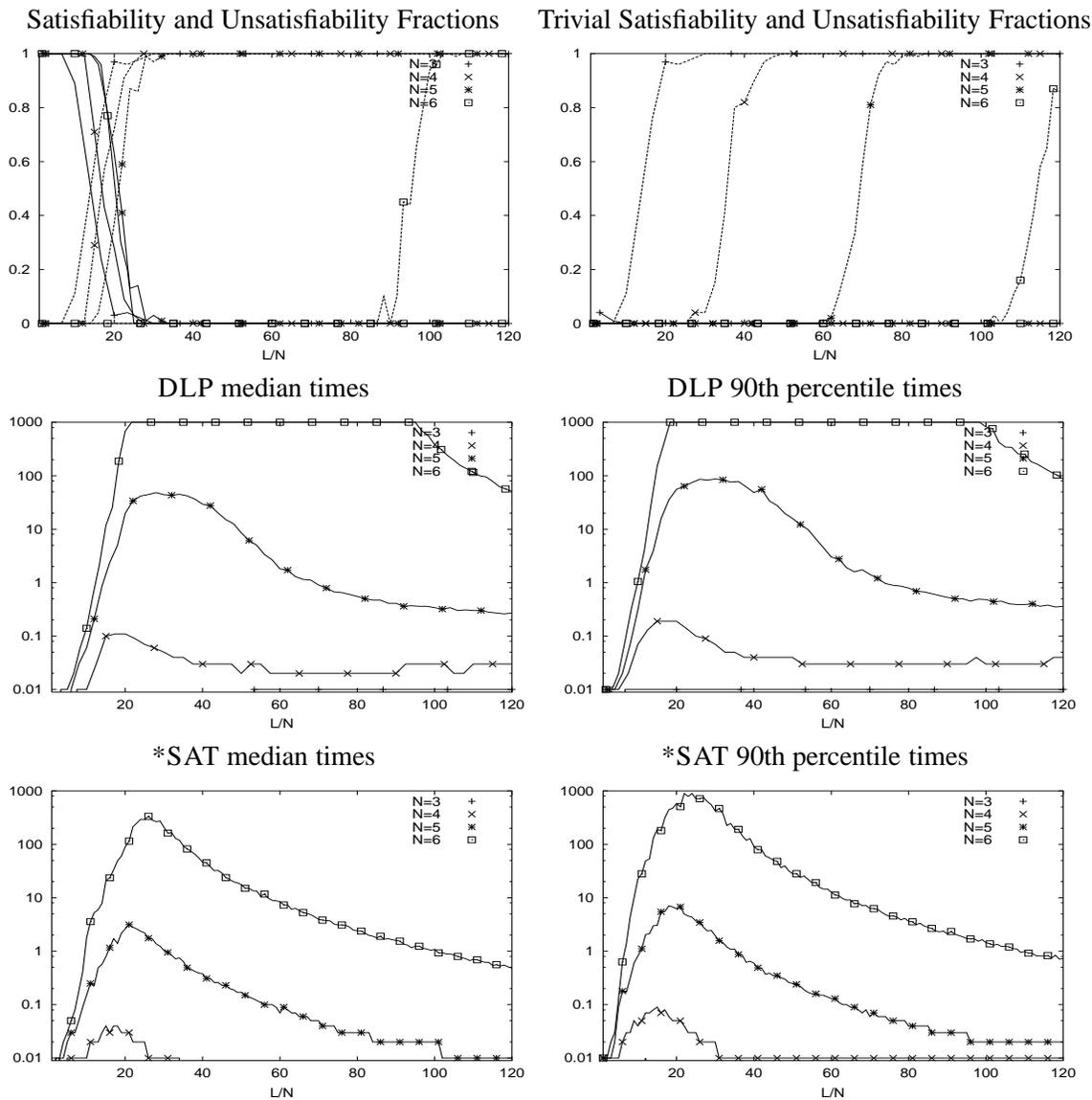

Figure 3: Results for $C = 3$, $m = 1$, $d = 1$, and $p = 0$ (either method).

The two methods with $p = 0.5$ are relatively close in maximum difficulty, with our new method generating somewhat harder formulae. However, our method produces difficult formulae, for both DLPand *SAT, over a much broader range of $L/N$ than does the original method.

Changing to $p = 0$ results in formulae that are orders of magnitude harder. This is not good, previous arguments to the contrary notwithstanding, as we would like to have a significant number of reasonable test sets to work with, and $p = 0$ allows only consideration of a very few values for $N$ before the formulae are totally impossible to solve with current systems, resulting in very few reasonable test sets.

So, at a maximum modal depth of $d = 1$ our method results in formulae that are of similar difficulty to the previously-generated formulae and still have trivially unsatisfiable formulae, but ones that do not seriously affect the difficulty of the test sets.

### 4.1.2 MODAL DEPTH 2

Restricting attention to a maximum modal depth of $d = 1$ is not very useful. Formulae with maximum modal depth of 1 are not representative of modal formulae in general, particularly as they have no nested modal operators. Sticking to a maximum modal depth of 1 seriously limits the significance of the generated tests.

We would thus like to be able to perform interesting experiments with larger maximum modal depths. So we performed a set of experiments with a maximum modal depth of $d = 2$. We started with a set of tests that corresponds to previously-performed experiments.

At depth $d = 2$, in the old method for $p = 0.5$ the time curves are dominated by a "half-dome" shape, whose steep side shows up where the number of trivially unsatisfiable formulae becomes large before the formulae become otherwise easy to solve, as shown in Figure 4. In fact, nearly all the unsatisfiable formulae here are trivially unsatisfiable.

This is an extremely serious flaw, as the difficulty of the test set is being drastically affected by these trivially unsatisfiable formulae. Changing to $p = 0$ is not a viable solution because at depth $d = 2$ such formulae are much too difficult to solve, as shown in Figure 5, where the median percentile exceeds the timeout before any formulae can be determined to be unsatisfiable, even for 3 propositional variables.

With our new method, as shown in Figure 6, the formulae are much more difficult to solve than the old method, because there is no abrupt drop-off from propositional unsatisfiability, but they are much easier to solve than those generated with $p = 0$. Further, trivially unsatisfiable formulae do not appear at all in the interesting portion of the test sets.

Nevertheless this choice of parameters ($d = 2$, $p = 0.5$) is not entirely suitable. The formulae are becoming too hard much too early. In particular, there are no unsatisfiable formulae that can be solved for $N > 3$, and thus the unsatisfiability plots cannot be distinguished from the x axis (recall Footnote 6). However, our new method does provide some advantages already, providing an interesting new set of tests, albeit one of limited size.

## 4.2 Increasing $p$

We would like to be able to produce better test sets for depth $d = 2$ and greater. One way of doing this is to increase the propositional probability $p$ from 0.5 to something like 0.6, increasing the number of propositional atoms and thus decreasing the difficulty of the generated formulae. This would be very problematic with previous generation methods as it would result in the trivially unsatisfiable formulae determining the results for even smaller numbers of clauses $L$, but with our method here it is not much of a problem.

To investigate the increasing of the the propositional probability, we ran a collection of tests with maximum modal depth $d = 2$ and propositional probability $p = 0.6$ with both the old method and our new method. The results of these tests are given in Figures 7 and 8. As before, the asymmetries between the satisfiability and unsatisfiability curves in Figure 8 for $N = 5, 6$ are due to the fact that many tests are not solved by DLP within the time limit (c.f., Footnote 6).

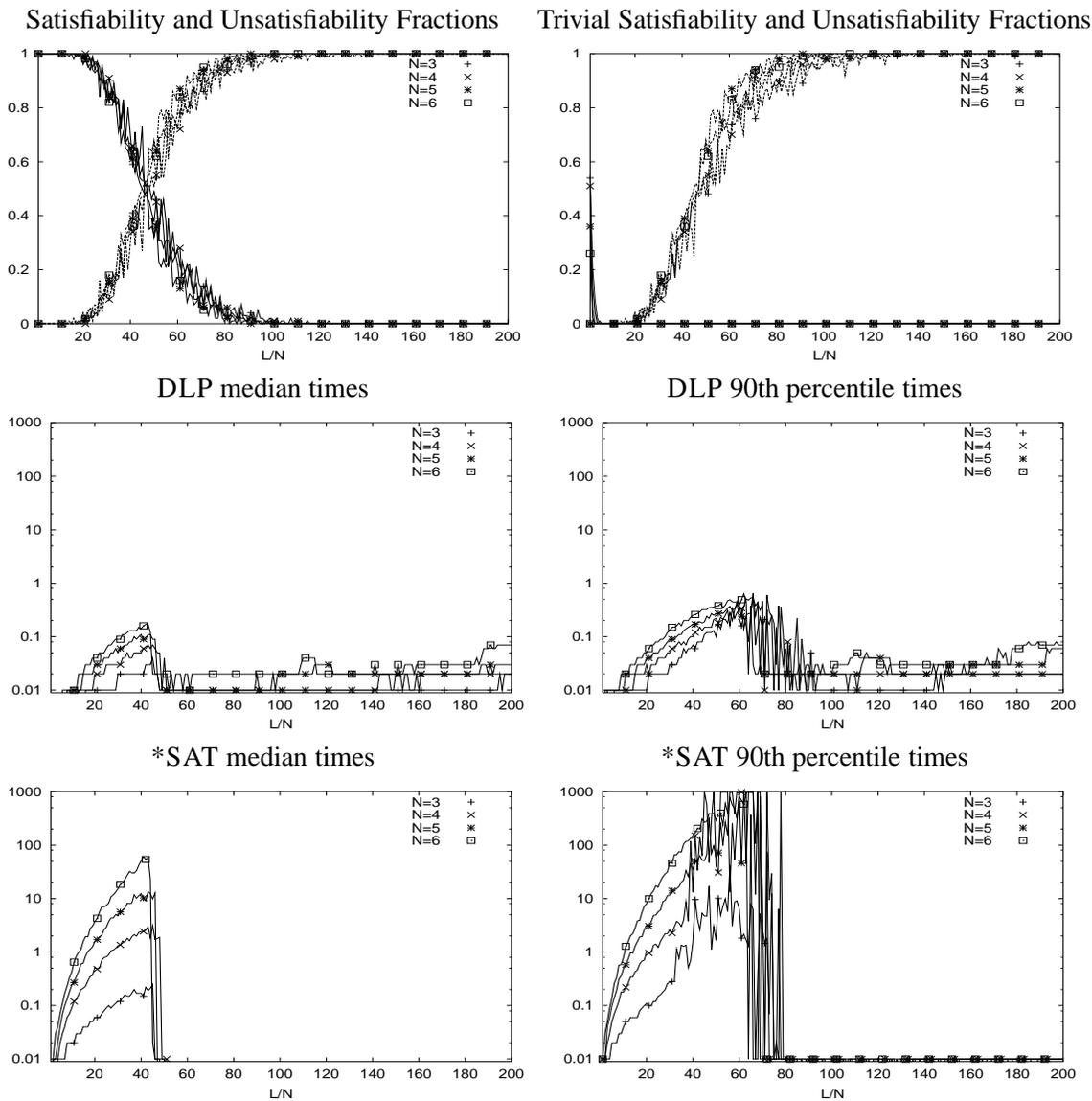

Figure 4: Results for $C = 3$, $m = 1$, $d = 2$, and $p = 0.5$ (old method)

As expected, the old method produces large numbers of trivially unsatisfiable formulae. These trivially unsatisfiable formulae show up much earlier than with $p = 0.5$, making the tests considerably easier, especially for *SAT.

Our new method produces hard formulae, but ones that are quite a bit easier than for $p = 0.5$. In particular, DLP solved all instances within the time limit for $N = 4$. Trivially unsatisfiable formulae do show up, but only well after the formulae are already unsatisfiable, and they do not significantly affect the difficulty of the tests.

So our method allows the creation of more-interesting tests at modal depths greater than 1, simply by adjusting $p$ to a value where the level of difficulty is appropriate. Trivial unsatisfiability is not a problem, whereas in the old method it was the most important feature of the test.

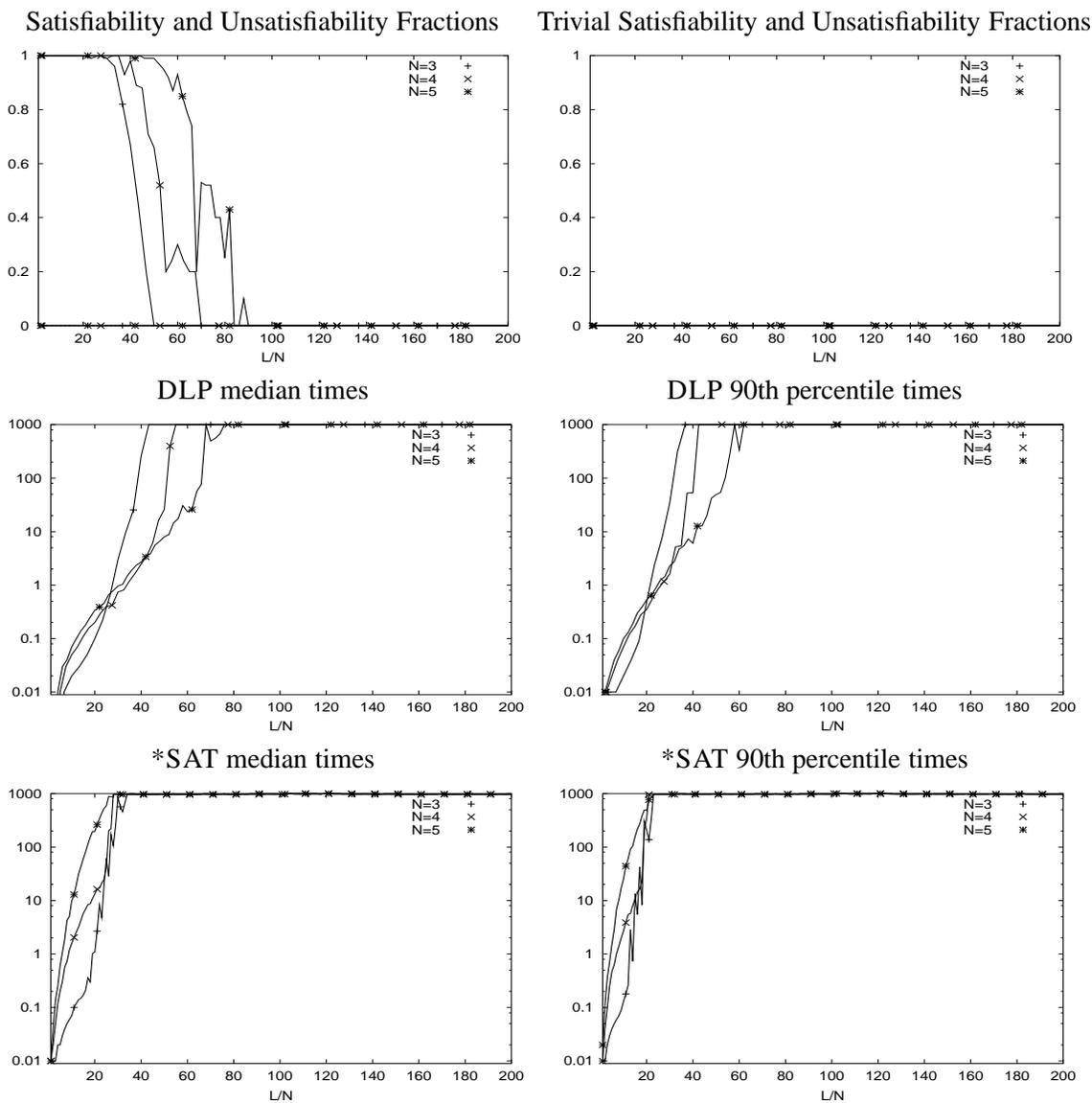

Figure 5: Results for $C = 3$, $m = 1$, $d = 2$, and $p = 0$ (either method)

## 4.3 Changing the Size of Clauses

A problem with increasing the propositional probability is that formulae become "too propositional" —that is, the source of difficulty becomes more and more the propositional component of the problem, and not the modal component. As we are interested in *modal* decision procedures, we do not want the main (or only) source of difficulty to be propositional reasoning.

We decided, therefore, to investigate a different method for modifying the difficulty of the generated formulae. We instead allow the number of literals in a clause $C$ to vary in a manner similar to the number of propositional atoms. If $C$ is an integer then each clause has that many literals. Otherwise, we allow either $\lfloor C \rfloor$ or $\lceil C \rceil$ literals in each clause, with probability $\lceil C \rceil - C$ and $C - \lfloor C \rfloor$, respectively. We then determine the number of propositional atoms in each clause based on the number of literals in that clause.

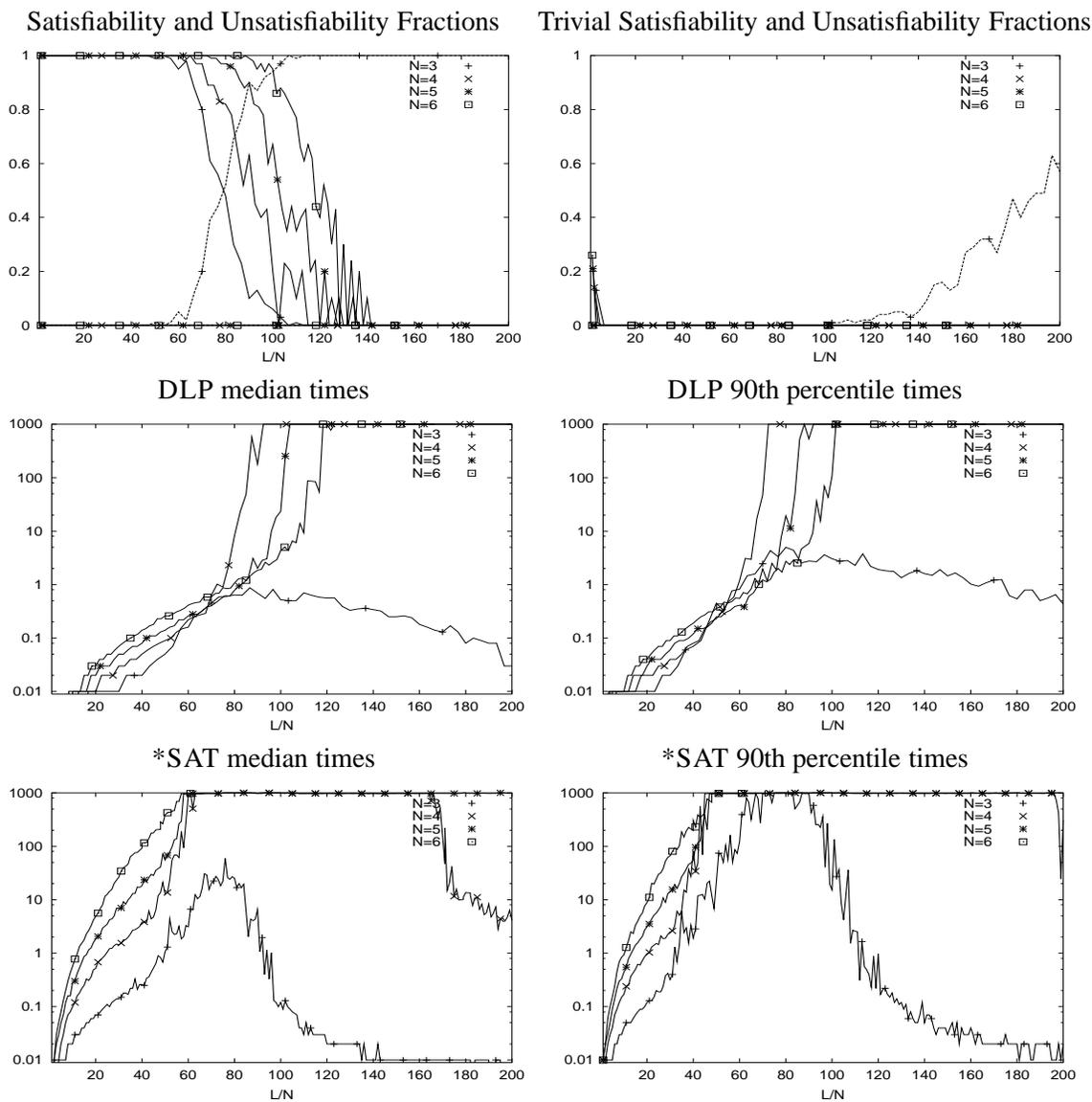

Figure 6: Results for $C = 3$, $m = 1$, $d = 2$, and $p = 0.5$ (our new method)

We generated $\text{CNF}_{\square_m}$ formulae with $C = 2.5$, $m = 1$, $d = 1$, and $p = 0.5$. The change from $C = 3$ to $C = 2.5$ produces fewer disjunctive choices and should result in easier formulae. The results of these tests are given in Figure 9.

These formulae are much easier than those generated with $C = 3$, although they are still quite hard and form a reasonable source of testing data. Trivially unsatisfiable formulae appear in large numbers only well after the formulae are all unsatisfiable and relatively easy.

To further illustrate the reduction in difficulty with smaller values of $C$ we generated formulae using $C = 2.25$, $m = 1$, $d = 1$, and $p = 0.5$. As shown in Figure 10, these formulae are even easier than for $C = 2.5$. Trivially unsatisfiable formulae do appear, but again only after the formulae become all unsatisfiable, and not until the formulae become easy, particularly for *SAT.

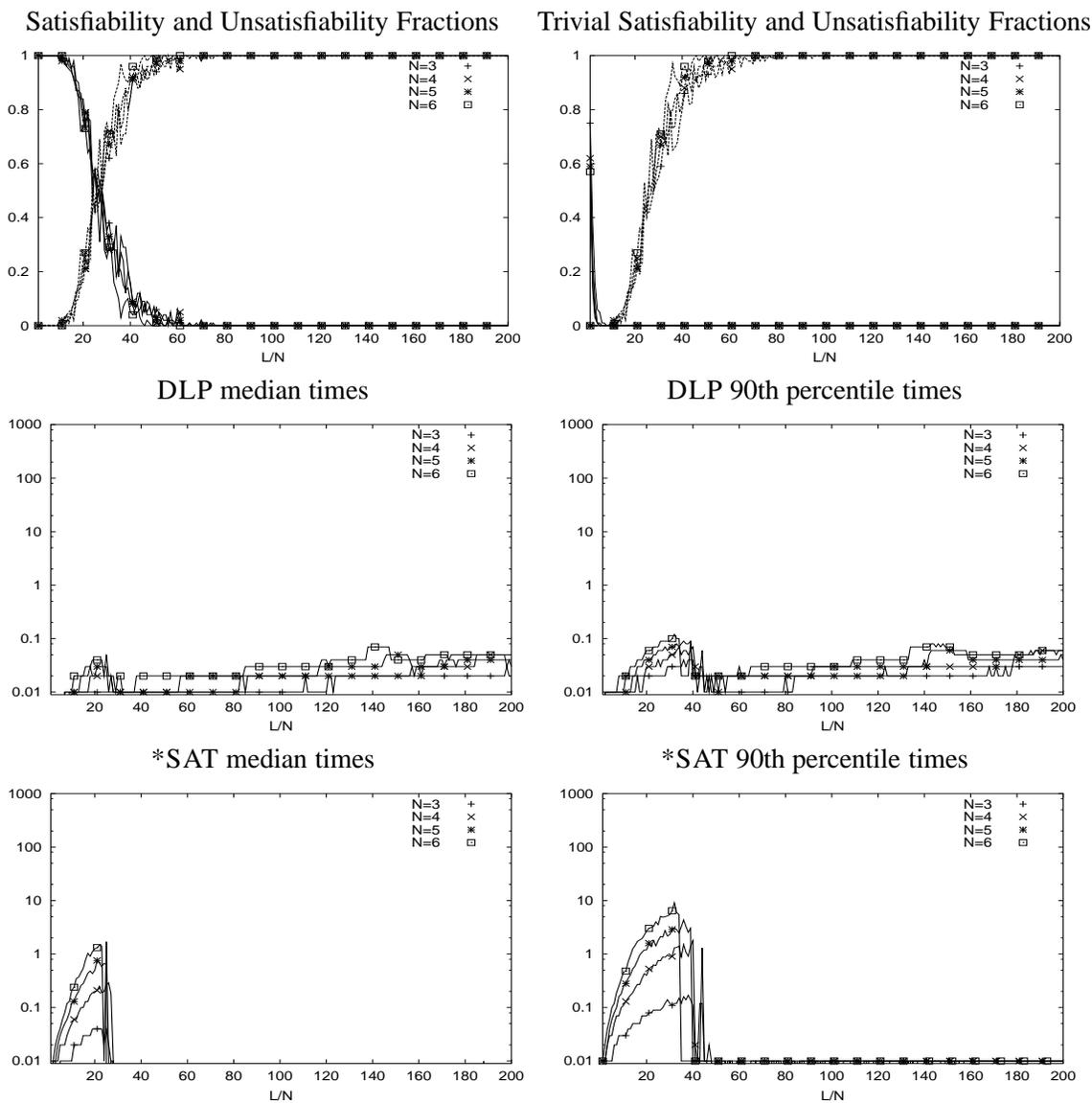

Figure 7: Results for $C = 3$, $m = 1$, $d = 2$, and $p = 0.6$ (old method)

At $C = 2.25$ we now have a reasonable set of formulae for maximum modal depth $d = 2$. With a maximum modal depth of 2, the formulae are much more representative than formulae with maximum modal depth of 1. The formulae are neither too easy nor too hard for current modal decision procedures so the satisfiability transition can be investigated for significant numbers of propositional variables.

Further, with this new method we can provide a collection of test sets that vary in difficulty by varying $C$. Most previous comparative test sets varied $N$, which is problematic because most interesting parameter sets become too hard for small values of $N$, in the range of 6 to 10.

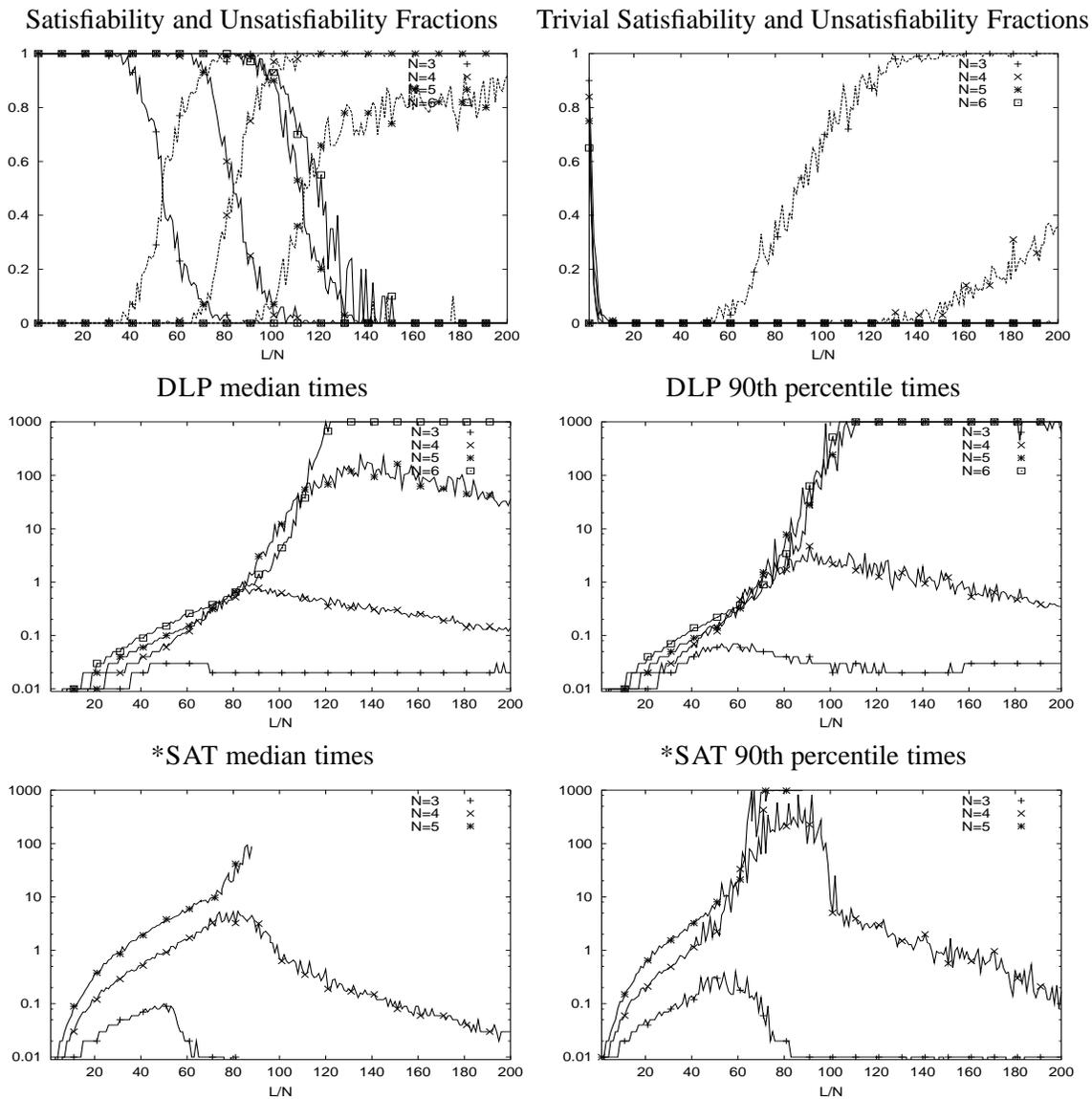

Figure 8: Results for $C = 3$, $m = 1$, $d = 2$, and $p = 0.6$ (our new method)

To illustrate the effects of varying $C$ we generated formulae using $N = 4$, $m = 1$, $d = 1$, and $p = 0.5$, varying $C$ from 2.2 to 2.8. As shown in Figure 11, this produces an interesting set of tests. The difficulty levels can be set appropriately. Trivially unsatisfiable formulae do appear, but only after the formulae become unsatisfiable anyway. Trivially unsatisfiable formulae do not influence the difficulty of the test.

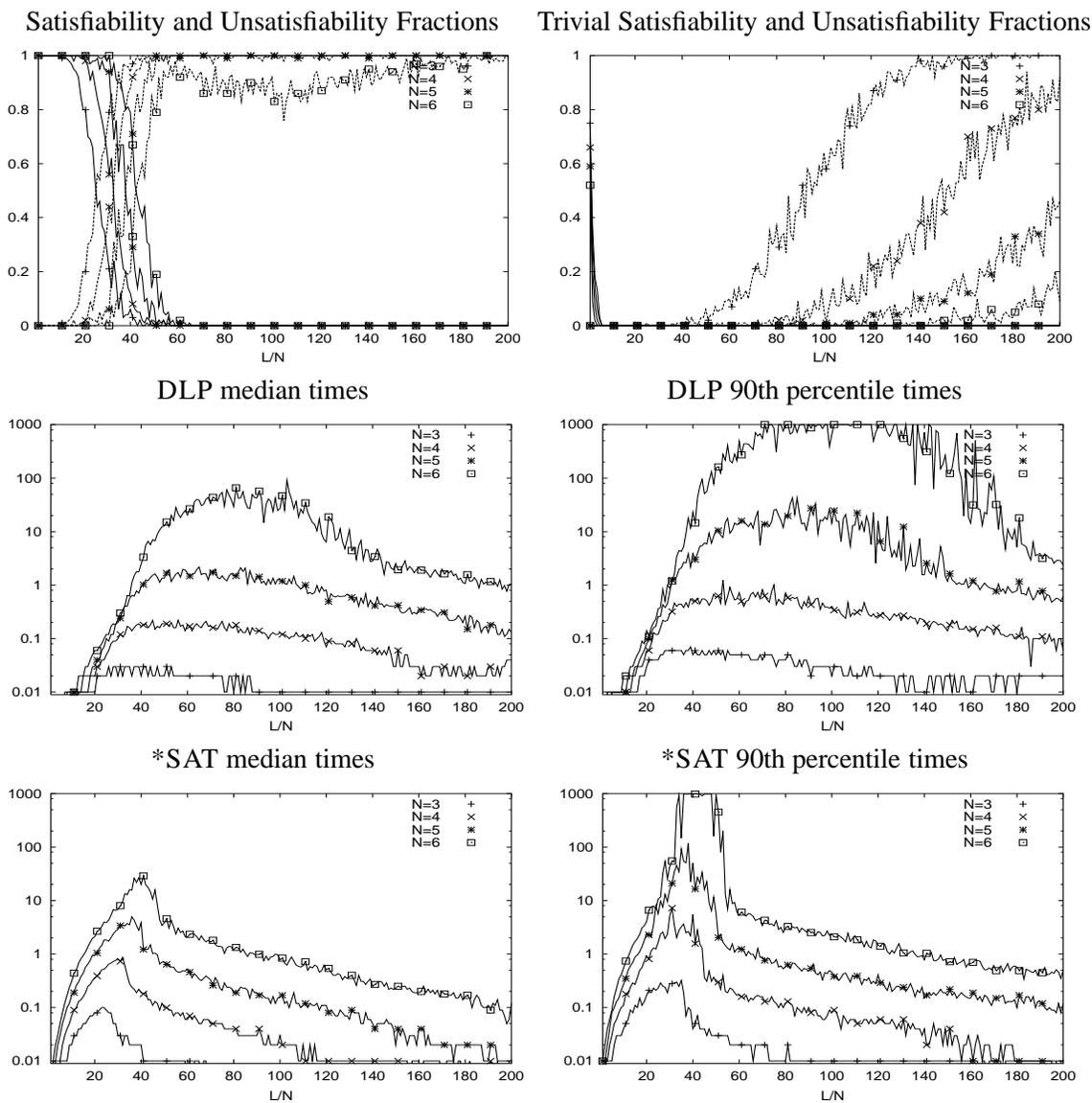

Figure 9: Results for $C = 2.5$, $m = 1$, $d = 2$, and $p = 0.5$ (our new method)

### 4.3.1 Modal Depth 3

Our method can be used to generate interesting test sets with modal depth $d = 3$. This depth is not at all interesting with previous methods—either the formulae are immensely difficult, such as for $p = 0$, or the behavior is dominated by trivial unsatisfiability, such as for $p = 0.5$.

For interesting levels of difficulty, we do have to reduce $C$ to values below 2.5. If $C$ is much larger, the formulae are too hard. However, with $C \leq 2.5$ we can produce interesting test sets, as shown in Figure 12. (The relevant asymmetry between the satisfiable and unsatisfiable rates curves for $N \geq 5$ is due to the high amount of tests exceeding the time limit.) Here the problems are hard even for $N \leq 5$ but doable, and there are no problems with trivially (un)satisfiable formulas.

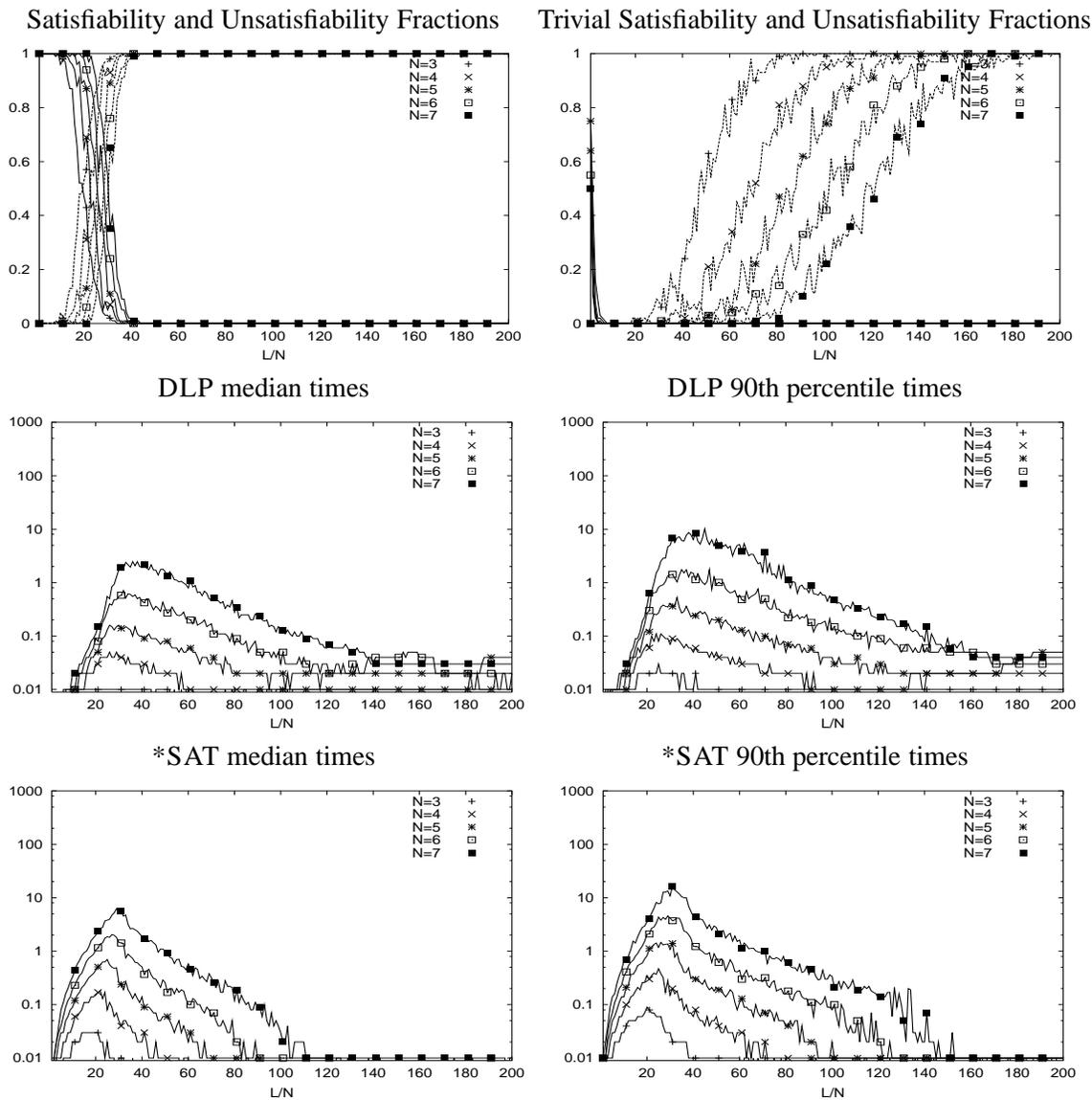

Figure 10: Results for $C = 2.25$, $m = 1$, $d = 2$, and $p = 0.5$ (our new method)

Our method now allows us fine control of the difficulty of tests. To make a test easier, we can just reduce the size of clauses by reducing the value(s) of $C$, or increase the propositional probability $p$. This control was missing with the previous method, as $C$ was restricted to integral value, and, anyway, was always set to 3 and making $p$ much different from 0.0 resulted in problems with trivial unsatisfiability for maximum modal depths greater than 1.

## 5. A New CNF$_{\Box_m}$ Generation Method: Advanced Version

Actually, our generator is much more general than what we have described so far. We allow direct specification of the probability distribution of the number of propositional atoms in a clause, and allow the distribution to be different for each modal depth from the top level to $d - 1$. We also allow direct specification of the probability distribution for the number of literals in a clause at each modal depth. Thus, the probability distribution for the number of propositional atoms depends on both the modal depth and the number of literals in the clause.

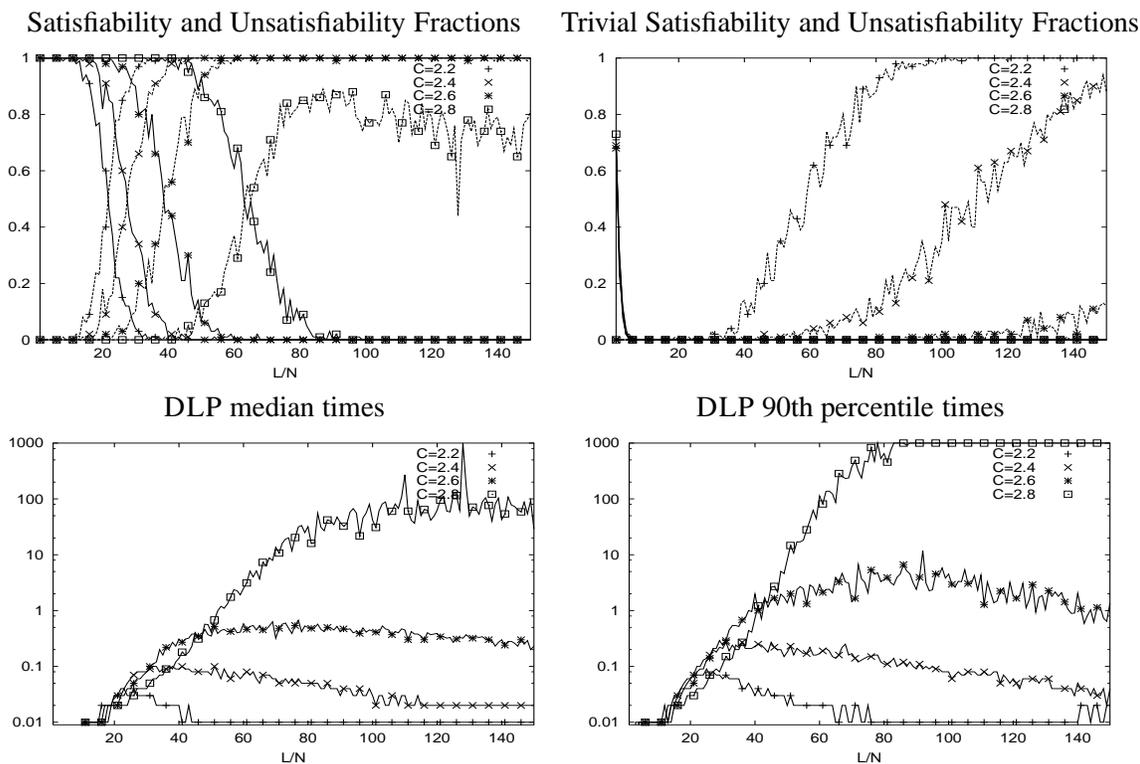

Figure 11: Results for $N = 4$, $m = 1$, $d = 2$, and $p = 0.5$ (our new method)

## 5.1 Generalization: Shaping the Probability Distributions.

The generator has two parameters to control the shape of formulae. The first parameter, $C$, is a list of lists (e.g., `[[0,0,1]]`) telling it how many disjuncts to put in each disjunction at each modal level. Each internal list represents a finite discrete probability distribution. For instance, the "`[0,0,1]`" says "0/1 of the disjunctions have 1 disjunct, 0/1 have 2 disjuncts, and 1/1 have 3 disjunctions" (fixed length 3). Because there is only one element of the list, this frequency is used at each modal depth, until the last. Other possibilities are, e.g., `[[1,1,1,1]]` (maximum length 4 with uniform distribution), `[[16,8,4,2,1]]` (maximum length 5 with exponential distribution), and so on.

The second parameter, $p$, is a list of lists of lists (e.g., `[[[],[],[0,3,3,0]]]`) that controls the propositional/modal rate. The top-level elements are for each modal depth (here all the same). The second-level elements are for disjunctions with 1,2,3,... disjunctions (here only the third matters as all disjunctions have three disjuncts). For instance, the "`[0,3,3,0]`" says "0/6 of the disjunctions have 0 propositional atoms, 3/6 have 1 propositional atom, 3/6 have 2 propositional atoms, and 0/6 have 3 propositional atoms" (that is, our new scheme discussed in the paper with $p = 0.5$; the old scheme with $p = 0.5$ is represented by `[[[],[],[1,3,3,1]]]`). Notice that the first element of the distributions in $C$ represents the value 1, whilst the first element of the distributions in $p$ represents the value 0. Setting the last element of each distribution to zero `[...,0]` eliminates all strictly propositional clauses, which are the main cause of trivial unsatisfiability; this is the way we implement the constraint $p \leq (C-1)/C$ of Section 4.1.

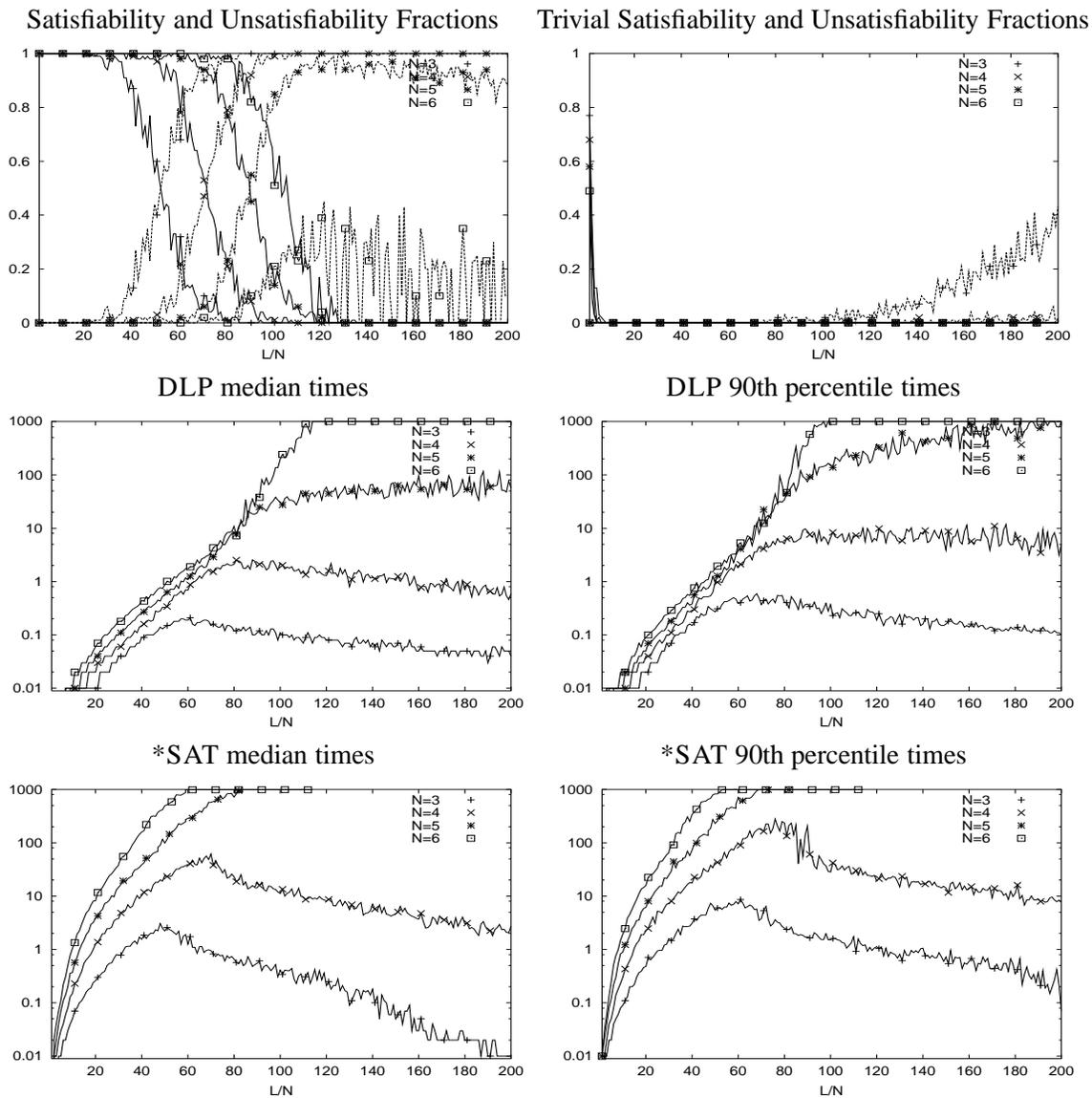

Figure 12: Results for $C = 2.25$, $m = 1$, $d = 3$, and $p = 0.5$

```
1   function rnd_CNF□m(d,m,L,N,p,C)
2       for i := 1 to L do                    /* generate L distinct random clauses */
3           repeat
4               Cli := rnd_clause(d,m,N,p,C);
5           until is_new(Cli);                 /* discards Cl if it already occurs */
6       return ⋀Li=1 Cli;

7   function rnd_clause(d,m,N,p,C)
8       K := rnd_length(d,C);                  /* select randomly the clause length */
9       P := rnd_propnum(d,p,K);               /* select randomly the prop/modal rate */
10      repeat
11          for j := 1 to P do                 /* generate P distinct random prop. literals */
12              lj := rnd_sign()·rnd_atom(0,m,N,p,C);
13          for j := P+1 to K do               /* generate K-P distinct random modal literals */
14              lj := rnd_sign()·rnd_atom(d,m,N,p,C);
15          Cl := ⋁Kj=1 lj;
16      until no_repeated_atoms_in(Cl);        /* discards Cl if contains repeated atoms */
17      return Sort(Cl);

18  function rnd_atom(d,m,N,p,C)
19      if d=0
20      then return rnd_propositional_atom(N); /* select randomly a prop. atom */
21      else
22          □r := rand_box(m);                 /* select randomly an indexed box */
23          Cl := rand_clause(d-1,m,N,p,C);
24          return □rCl ;
```

Figure 13: Schema of the new CNF$_{□_m}$ random generator.

For instance, the plots of Figures 1-12 can be obtained with the following choices of C and p:

| Fig. | C | p | C (advanced version) | p (advanced version) |
|------|-----|-----------|---------------------------------|---------------------------------|
| 1, 4 | 3 | 0.5 (old) | `[[0,0,1]]` | `[[[],[],[1,3,3,1]]` |
| 2, 6 | 3 | 0.5 (new) | `[[0,0,1]]` | `[[[],[],[0,3,3,0]]` |
| 3, 5 | 3 | 0 | `[[0,0,1]]` | `[[[],[],[1,0,0,0]]` |
| 7 | 3 | 0.6 (old) | `[[0,0,1]]` | `[[[],[],[8,36,54,27]]` |
| 8 | 3 | 0.6 (new) | `[[0,0,1]]` | `[[[],[],[0,1,4,0]]` |
| 9 | 2.5 | 0.5 (new) | `[[0,1,1]]` | `[[[],[0,3,0],[0,3,3,0]]` |
| 10, 12 | 2.25 | 0.5 (new) | `[[0,2,1]]` | `[[[],[0,3,0],[0,3,3,0]]` |
| 11 | 2.2, 2.4, | 0.5 (new) | `[[0,4,1]],[[0,3,2]]` | `[[[],[0,3,0],[0,3,3,0]]` |
| | 2.6, 2.8 | | `[[0,2,3]],[[0,1,4]]` | |

Our generator works as described in Figure 13. The function is_new(Cli) checks if $Cl_i \neq Cl_j$, $\forall \, j < i$; rnd_length(d,C) selects randomly the clause length according to the $d + 1$-th distribution in C (e.g, if d is 1 and C is `[[0,1,1]` `[1,2]` `[1]]`, it returns 1 with probability 1/3 and 2 with probability 2/3); *rnd_propnum(d,p,K)* selects randomly the number of propositional atoms per clause $P$ according to the $[d+1, K]$-th distribution in $p$ (e.g, if $d$ is 1, $K$ is 2 and $p$ is `[[[],[0,1,0],[0,1,0,0]] [[1,0]`<u>`[0,1,0]`</u>`]]`, it returns 1 deterministically); *rnd_sign* selects randomly either the positive or negative sign with equal probability; *no_repeated_atoms_in(Cl)* checks if the clause $Cl$ contains no repeated atom; *Sort(Cl)* returns the clause $Cl$ sorted according to some criterium; *rnd_propositional_atom(N)* selects with uniform probability one of the $N$ propositional atoms $A_i$; *rnd_box(m)* selects with uniform probability one of the $m$ indexed boxes $\Box_r$.

When eliminating duplicated atoms in a clause, we take care not to disturb these probabilities by first determining the "shape" of a clause (rows 8-9 in Figure 13), and only then instantiating that with propositional variables (rows 10-16 in Figure 13). If a clause has repeated atoms, either propositional or modal, the instantiation is rejected and another instantiation of the shape is performed. If we did not take care in this way we would generate too few "small" atoms because there are fewer small atoms than large atoms, resulting in a greater chance of rejecting small atoms because of repetition.

The elimination of duplicated atoms in a clause is not only a matter of elimination of redundancies, but also of elimination of a source of flaws. In fact, one might generate top-level clauses like $... \wedge (\neg\Box_1(A_1 \vee \neg A_1) \vee \neg\Box_2(A_2 \vee \neg A_2)) \wedge ...$, which would make the whole formula inconsistent.

**Example 5.1** We try to guess a parameter set by which the new random generator can potentially generate the following $\text{CNF}_{\Box_m}$ formula $\varphi$:

$$
\begin{array}{ll}
(\quad \neg A_3 \vee \Box_1(\neg A_4 \vee \neg\Box_1 A_1) \vee \Box_1(\neg A_1 \vee \neg\Box_1 A_2) \quad) \wedge \\
(\quad \neg A_1 \vee \Box_1(A_3 \vee \neg\Box_1 A_2) \vee \neg\Box_1(\Box_1 \neg A_4) \quad) \wedge \\
(\quad \neg A_4 \vee \neg\Box_1(A_2 \vee \Box_1 \neg A_1) \quad) \wedge \\
(\quad A_1 \vee \neg\Box_1(\neg\Box_1 A_4) \quad).
\end{array}
\tag{2}
$$

After a quick look we set $m = 1$, $d = 2$, $N = 4$, $L = 4$. At top level we have 0 unary, 2 binary and 2 ternary clauses; at depth 1 we have 2 unary and 4 binary clauses; at depth 2 we have only 6 unary clauses. Thus, we can set

$$
\texttt{C = [[0,2,2],[2,4],[6]].}
\tag{3}
$$

At top level there are no unary clauses (we represent this fact by the empty list "`[]`"), the 2 binary clauses have 1 propositional literal, and the 2 ternary clauses have 1 propositional literal; at depth 1, the 2 unary clauses have 0 propositional literals, while the 4 binary clauses have 1 propositional literal. (There is no need to provide any information for depth 2, as all clauses are purely propositional.) Thus, we can set

$$
\texttt{p = [[[],[0,2,0],[0,2,0,0]] [[2,0],[0,4,0]]].}
\tag{4}
$$

The two expressions can then be normalized into:

$$
\begin{array}{l}
\texttt{C = [[0,1,1],[1,2],[1]]} \\
\texttt{p = [[[],[0,1,0],[0,1,0,0]] [[1,0],[0,1,0]]].}
\end{array}
\tag{5}
$$

Notice that any other setting of $C$, $p$ obtained by changing the non-zero values in (5) into other non-zero values, or turning zeros into non-zeros (but not vice versa!), will do the work, just with a different probability. For instance, turning the first list in $C$ into `[1,1,1]` allows for generating also unary clauses at top level; anyway, with probability $(2/3)^L$ the generator may still produce formulae with only binary and ternary clauses at top level. □

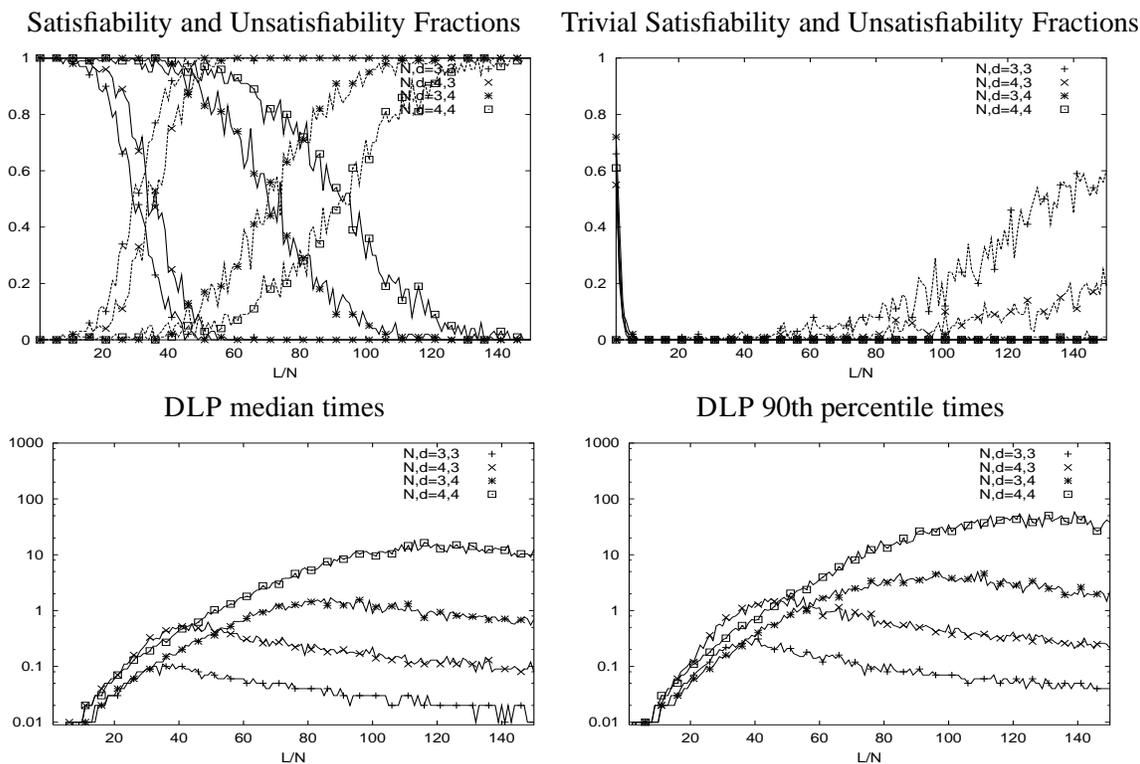

Figure 14: Results for DLP with $d = 3, 4$, $N = 3, 4$, $C = $ `[[1,8,1]]`,
$p = $ `[[[1,0],[0,1,0],[0,1,1,0]]]`.

As an illustration of our general method, we present a set of tests with $m = 1$, $d = 3, 4$, $N = 3, 4,$, $C = $ `[[1,8,1]]`, and $p = $ `[[[1,0],[0,1,0],[0,1,1,0]]]`. This set of tests introduces a small fraction of single-literal clauses that contain a modal literal (except at the greatest modal depth, where they contain, of course, a single propositional literal). The results of tests are given in Figure 14. Again, trivial instances occur only out the interesting zone. Here we can generate interesting test sets even with modal depth $4$.

## 5.2 Varying the Probability Distributions with the Depth

Our new method provides the ability to fine-tune the distribution of both the size and the propositional/modal rate of the clauses *at every depth*. This fine tuning results in a very large number of parameters, and so far in this paper we have only investigated distributions that conform to the scheme described above or ones that correspond to the 3CNF$_{\Box_m}$ generation method previously used.

To give an example of how to vary the probability distributions with the nesting depth of the clauses, we consider the case with $d = 4, 5$, $m = 1$, $N = 3, 4, 5$, $C = $ `[[1,8,1],[1,2]]`, $p = $ `[[[1,0],[0,1,0],[0,1,1,0]],[[1,0],[0,1,0]]]`. The results of the tests are given in Figure 15.

The $C$ parameter says that the probability distributions of the length of the clauses occurring at nesting depth 0 and $\geq 1$ are `[1,8,1]` and `[1,2]` respectively. (When not explicitly specified, it is

374

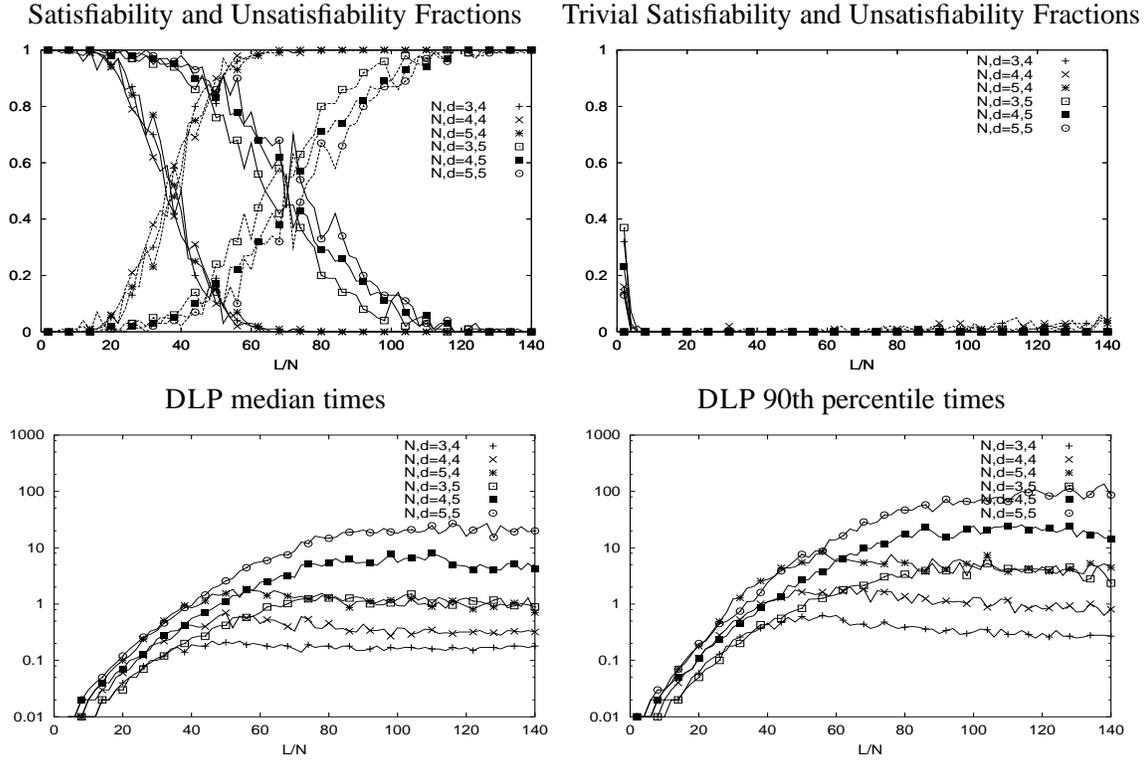

Figure 15: Results for DLP with $d = 4, 5$, $m = 1$, $N = 3, 4, 5$, $C = $`[[1,8,1],[1,2]]`, $p = $`[[[1,0],[0,1,0],[0,1,1,0]],[[1,0],[0,1,0]]]`.

considered the last distribution by default, as in the case of depth $> 1$.) Thus, the top-level clauses are on average 1/10 unary, 8/10 binary 1/10 ternary, while the clauses occurring at depth $\geq 1$ are on average 1/3 unary and 2/3 binary.

The $p$ parameter says that the lists of probability distributions of the propositional/modal ratio at nesting depth 0 and $\geq 1$ are `[[1,0],[0,1,0],[0,1,1,0]]` and `[[1,0],[0,1,0]]` respectively. Thus, at every depth, unary clauses have no propositional literal and binary clauses have 1 propositional and 1 modal literal. The top-level ternary clauses have either 1 or 2 propositional literals, with equal probability.

Notice that at top level the distributions are identical to those of Figure 14, whilst at depth $\geq 1$ there are no more ternary clauses and a higher fraction of unary clauses. These slight modifications allow reasonable test sets with $d = 5$ and $N = 5$. Moreover, trivial instances have nearly disappeared.

## 6. Generality of the Method

We have already observed (Horrocks et al., 2000) that for normal modal logics, from $\mathbf{K_{(m)}}$ upward, there is no loss in the restriction to $\text{CNF}_{\Box_m}$ formulae, as there is an equivalence between arbitrary normal modal formulae and $CNF_{\Box_m}$ formulae[9]. We may wonder how well our generation technique covers the whole space of $CNF_{\Box_m}$ formulae, and how well we can approximate a restricted subclass of this space. Example 5.1 represents an instance of a very general property of our random generation technique, which we present and discuss below.

Now we assume that the $rnd\_CNF_{\Box_m}$ of Figure 13 is a "purely random" generator, i.e., it performs all non-deterministic choices independently and in a pure random way. (Of course pseudo-random generators only approximate this feature.) Moreover, with no loss of generality, we restrict our discussion to $CNF_{\Box_m}$ formulae which have no repeated clauses at top level and no repeated atoms inside any clause at any level, and in which atoms are *sorted* within each clause, according to the generic function *Sort()* of Figure 13. The former allows for considering only formulae which are already simplified out; the latter allows for considering only one representative for each class of formulae which are equivalent modulo order permutations. As discussed by Giunchiglia et al. (2000), the latter allows for further simplifying subformulae like, e.g., $\Box(A_1 \lor A_2) \lor \Box(A_2 \lor A_1)$ or $\Box(A_1 \lor A_2) \lor \neg\Box(A_2 \lor A_1)$.

Let $\varphi$ be a sorted $CNF_{\Box_m}$ formula of depth $d$ and with $L$ top-level clauses built on all the propositional atoms $\{A_1, ... A_N\}$ and on all the modal boxes $\{\Box_1, ... \Box_m\}$, which has no repeated clause at top level and no repeated atoms inside any clause at any level. Then we can construct $C$ and $p$ so that, for each $i$, $j$, $r$:

(*a*) the $j$-th element of the $i$-th sublist in $C$ is non-zero if and only if there is a clause of length $j$ occurring at depth $i$ in $\varphi$, and

(*b*) the $r + 1$-the element of the $j$-th sub-sublist of the $i$-th sublist in $p$ is non-zero if and only if there is a clause of length $j$ occurring at depth $i$ which contains $r$ propositional literals.

One possible operative technique to build $C$ and $p$ works as follows. Initialize $C$ as a list of $d + 1$ sublists. Then, for every depth level $i \in \{0, ..., d\}$, set the $i$-th sublist of $C$ as follows:

(i) set the size $K$ of the sublist as the maximum size of clauses occurring in $\varphi$ at depth $i$;

(ii) for all $j \in \{1, .., K\}$, count the number of clauses of length $j$ occurring in $\varphi$ at depth $i$, and append the result to the sublist.

Initialize $p$ as a list of $d$ sublists of sub-sublists. Then, for every depth level $i \in \{0, ..., d - 1\}$, set the $i$-th sublist of $p$ as follows:

(i) look at $C$: set the size $K$ of the sublist as the maximum size of clauses occurring in $\varphi$ at depth $i$;

(ii) for all $j \in \{1, .., K\}$, generate the $j$-th sub-sublist as follows:

- look at $C$: if the number of clauses of length $j$ occurring at depth $i$ is non-zero, then set the length $l$ of the sub-sublist to $j + 1$, else set $l$ to 0;

---

9. This holds for all modal normal logics from $\mathbf{K_{(m)}}$ upward, as the conversion works recursively on the depth of the formula, from the leaves to the root, each time applying to sub-formulae the propositional CNF conversion and the transformation

$$\Box_r \bigwedge_j \bigvee_i \varphi_{ij} \Longrightarrow \bigwedge_j \Box_r \bigvee_i \varphi_{ij},$$

which preserves validity in such logics.

- for all $r \in \{0, .., l-1\}$, count the number of clauses of length $j$ occurring in $\varphi$ at depth $i$ which have $r$ propositional literals, and append the result to the sub-sublist.

Example 5.1 represents an instance of application of the above technique for construction $C$ and $p$ from $\varphi$. Notice that the $C$ and $p$ parameters not only verify points $(a)$ and $(b)$ above, but are such that the probability distributions mimic the actual number of occurrences of the different kinds of clauses.

**Theorem 6.1** *Let rnd_CNF$_{\Box_m}$ be a purely random generator as in Figure 13. Let $\varphi$ be a sorted CNF$_{\Box_m}$ formula of depth $d$ and with $L$ top-level clauses built on all the propositional atoms in $\{A_1, ...A_N\}$ and on all the modal boxes in $\{\Box_1, ...\Box_m\}$, which has no repeated clause at top level and no repeated atoms inside any clause at any level. Let $C$ and $p$ be built from $\varphi$ so that to verify points $(a)$ and $(b)$ above. Let $C'$ and $p'$ be obtained from $C$ and $p$ respectively by substituting some zero-values with some non-zero values. Then we have:*

(i) *rnd_CNF$_{\Box_m}$(d,m,L,N,p,C) returns $\varphi$ with some non-zero probability $\mathcal{P}$;*

(ii) *rnd_CNF$_{\Box_m}$(d,m,L,N,p',C') returns $\varphi$ with some non-zero probability $\mathcal{P}' \leq \mathcal{P}$.*

**Proof** The fully-detailed proof is reported in Appendix. Here we sketch the main steps.

The following facts come straightforwardly by induction on the structure of $\varphi$:

1. every propositional atom occurring in $\varphi$ at some depth $i$ is returned with the same non-zero probability $\mathcal{P}_1$ by both *rnd_atom(0,m,N,p,C)* and *rnd_atom(0,m,N,p',C')*;

2. every modal atom $\Box_v Cl$ occurring in $\varphi$ at some depth $i$ is returned with some non-zero probability $\mathcal{P}_2$ by *rnd_atom(d-i,m,N,p,C)*, and is returned with some non-zero probability $\mathcal{P}'_2 \leq \mathcal{P}_2$ by *rnd_atom(d-i,m,N,p',C')*;

3. every clause $Cl$ occurring in $\varphi$ at some depth $i$ is returned with some non-zero probability $\mathcal{P}_3$ by *rnd_clause(d-i,m,N,p,C)*, and is returned with some non-zero probability $\mathcal{P}'_3 \leq \mathcal{P}_3$ by *rnd_clause(d-i,m,N,p',C')*.

Thus, every top level clause $Cl_k$ is returned by *rnd_clause(d,m,N,p,C)* and *rnd_clause(d,m,N,p',C')* with some non-zero probabilities $\mathcal{P}_k$ and $\mathcal{P}'_k$ respectively, being $\mathcal{P}'_k \leq \mathcal{P}_k$. From this fact, it comes straightforwardly that $\varphi$ is returned by *rnd_CNF$_{\Box_m}$(d,m,L,N,p,C)* and *rnd_CNF$_{\Box_m}$(d,m,L,N,p',C')* with some non-zero probabilities $\mathcal{P}$ and $\mathcal{P}'$ respectively, being $\mathcal{P}' \leq \mathcal{P}$.    Q.E.D.

    Q.E.D.

From a theoretical viewpoint, Theorem 6.1 $(i)$ shows that our generation technique is very general, because, for every CNF$_{\Box_m}$ formula $\varphi$, there exists a choice for the parameters s.t. a purely random generator returns $\varphi$ with some non-zero probability $\mathcal{P}$.

Of course, the choice criterium for $C$ and $p$ suggested by points $(a)$ and $(b)$ is not unique as, for example, any other setting obtained from it by turning zeros into non-zeros would match the requirements. As an extreme case, we might think to do very general choices like

$$C = [[1,1,1,...],...] \quad p = [[[[1,1],[1,1,1],[1,1,1,1]...]]]. \quad (6)$$

which guarantee to have every possible CNF$_{\Box_m}$ formula within a given bound in clause size with non-zero probability. Anyway, Theorem 6.1 $(ii)$ shows that, extending the number of non-zeros values, the probability of generating $\varphi$ decreases.

For instance, consider Example 5.1. Turning the first list in $C$ of (5) into `[1,1,1]` would still allow for generating the formula (2), but it would allow for generating also unary clauses at top level with probability $1 - (2/3)^L$, which converges quickly to 1 with $L$.

Usually we are not interested in randomly generating one precise formula with some non-zero probability —which would be rather small anyway— but rather to randomly generate a class of formulae which are as similar as possible a given target class of formulae. Adding redundant non-zeros would extend the range of shapes for formulae, extending the variance and lowering the resemblance to the target class of formulae.

## 7. Discussion

### 7.1 The Basic and the Advanced Method

Our new testing method can be used at two different levels, depending on the attitude —and on the skills and experience— of the user.

In the *basic* usage the clause length $C$ is represented by lists with either only one non-zero element (e.g., `[[0,0,1]]`, meaning "clause length 2") or only two adjacent non-zero elements (e.g., `[[0,2,1]]`, meaning "clause length 2 or 3, with probability 2/3 and 1/3 respectively"); similarly, the propositional/modal rate $p$ is represented by lists with either only one non-zero element (e.g., `[[[],[],[0,1,0,0]]]`, meaning "1 propositional literal per clause") or only two non-zero adjacent elements (e.g., `[[[],[],[0,3,2,0]]]`, meaning "either 1 or 2 propositional literals per clause, with probability 3/5 and 2/5 respectively"); the distributions do not vary with the depth.

In the basic way the random generator is used as a "flawless"[10] extension of the $3CNF_{\square_m}$ method of Giunchiglia and Sebastiani (1996), which allows for setting the clause length to either fixed integer values or to non-integer average values. The number of parameters is kept relatively small, so that to allow a coarse-grained coverage of a significant subspace with an affordable number of tests.

In the *advanced* usage, it is possible to apply any finite probability distributions to both $C$ and $p$; moreover, it is possible to use different distributions at different depths. This opens a huge amount of possibilities, but requires some skills and experience from the user: the representation of sophisticated multi-level distributions may be rather complicated, and may thus require some practice; moreover, the usage of complex distributions requires some care, as the presence non-constant distributions in both clause length and propositional/modal rate may significantly enlarge the variance of the features of the generated formulae, making the effects of the tests more unpredictable and instance-dependent.

In order to guide the user, we provide some general suggestions for choosing the parameter sets in a testing session. They come from both theoretical issues and our practical experience in using the generator.

- Avoid generating purely propositional top-level clauses, that is, set `p = [[...,0],...]`. See Sections 4.1 and 5.1. If possible, avoid generating unary top-level clause, that is, set `C = [[0,...],...]`). See also Section 7.5.

---

10. In the sense of "free from the flaw highlighted in the work by Hustadt and Schmidt (1999) and Giunchiglia et al. (2000)".

- In organizing a testing session, fix the parameter sets according to the following order and directives.

  (i) Fix `d`. With `d=1` the search is mostly dominated by its propositional component, with `d>2` it tends to be dominated by its modal component. `d=2` is typically a good start.

  (ii) Fix `m`. `m` substantially partitions the problem into `m` independent problems. Increasing `m`, the samples tend to be more likely-satisfiable. `m=1` is typically a good start.

  (iii) Set `C`. Increasing the top level values of `C`, the samples tend to be more likely-satisfiable and the propositional component of search increases, so that the transition area moves to the right and the hardness peaks grow. Average values in $[2.0, 3.0]$ for the top level distributions of `C` are typically a good start.

  (iv) Set `p`. Decreasing the top level values of `p`, the modal component of search increases. For the the top level distributions of `p`, having on average half of top-level atoms propositional (that is, the $p = 0.5$ of Section 4) is typically a good start.

  (v) For each choice of the above parameters, increase `N`, starting from (at least) the maximum length in `C`, until the desired level of hardness is reached.

  (vi) Make `L` vary within the satisfiability transition area.

- When dealing with `C` and `p`, focus on top-level clause distributions first. Small variations of `C` and `p` at top level may cause big variations in hardness and satisfiability probability. Variations at lower levels typically cause much smaller effects.

- Use convex distributions: e.g., `[1,5,1]` and `[5,1,5]` have the same mean value, but the variance of the former is much smaller than that of the latter.

- Do keep `L` ranging in the satisfiability transition area: increasing `L` out of it, the fraction of trivially unsatisfiable samples can become relevant. To determine the satisfiability transition area, make a preliminary check with few samples per point (say, 10) using dichotomic search.

- Unlike `N` (and `m`), the parameters `d`, `C`, `p` make the formulas vary their shape. Thus, we suggest to group together plots with the same `d`, `C` and `p` values and increasing `N`'s.

On the whole, the large number of parameters makes it impossible to cover the parameter space in a reasonable amount of testing. However, just about any $\text{CNF}_{\square_m}$ formula shape can be generated so that the method described in Section 6 can be used to produce random formulae reasonably similar to some formula(e) of interest.

## 7.2 Comparison with the Old 3CNF$_{\square_m}$ Method

On the whole, the new method inherits all the features of the old 3CNF$_{\square_m}$ method.

**Scalability:** Increasing $N$, $d$ (and also the average clause length in $C$) the difficulty of the generated problems scales up at will. Thus it is possible to compare how the performance of different systems scale up with problems of increasing difficulty, for each source of difficulty (e.g., size, depth, etc.).

**Valid vs. not-valid balance:** The parameter $L$ allows for tuning the satisfiability rate of the formula at will. Moreover, it is always possible to choose $L$ to generate testbeds with about a 50%-satisfiable rate, which allows for the maximum uncertainty.

**Termination:** The new method allows for generating test sets of up to depth 3-4 which are run by state-of-the-art systems in a reasonable amount of time.

**Reproducibility:** The results of each testbed are easy to reproduce because the generator's code and all the parameters' values are made publicly available.

**Parameterization:** The random generation of $CNF_{\Box_m}$ formulae is fully parametric.

**Data organization:** The most natural way to use the new random generator is to generate tests and plot data by increasing values of one or two parameters. This allows for easy, quantitative and qualitative evaluations of the performances of the different procedures under test.

Moreover, the new method improves the $3CNF_{\Box_m}$ method for the following features.

**Representativeness:** As stated in Section 6, $CNF_{\Box_m}$ formulae represent all formulae in the normal modal logics from $\mathbf{K}_{(\mathbf{m})}$ upward, as there is an equivalence-preserving way of converting all modal formulae into $CNF_{\Box_m}$. From Theorem 6.1, the new method allows for a very fine-grained sampling of the class of $CNF_{\Box_m}$ formulae.

**Difficulty:** The random $CNF_{\Box_m}$ formulae with $d \geq 2$ and $N \geq 4$ provide challenging test sets for state-of-the art procedures. $CNF_{\Box_m}$ formulae with $d \geq 4$ and $N \geq 9$ can be well considered as challenges for next-generation systems. (Of course, it is not a problem to generate easy problems too.)

**Control:** The parameters $N$, $d$ and $C$ allow for controlling *monotonically* the difficulty of the test set. (E.g., if you increase $N$, you are reasonably sure that your mean/median CPU time plots will increase.) The parameter $L$ allows for controlling the satisfiability rate. Monotonicity allows for controlling one feature by simply increasing or decreasing one value, and thus for eliminating uninteresting areas of the input space.

**Modal vs. propositional balance:** The size of the Kripke models spanned by the decision procedures has increased exponentially with the higher modal depths reached by the new test sets; moreover, the probability of repeated top-level atoms has dramatically reduced.[11] Consequently, unlike the tests by Hustadt and Schmidt (1999) and Giunchiglia et al. (2000) the search is no longer dominated by the pure propositional component of reasoning, and the empirical results show that a large number of modal successors are explored.

Finally, the new method completely removes or drastically reduces the effects of the following problems.

**Redundancy:** Propositional and modal redundancy had already been eliminated in the last versions of the $3CNF_{\Box_m}$ method (Giunchiglia et al., 2000). Moreover, the new method allows for eliminating all strictly propositional clauses.

**Triviality:** The main cause of trivial unsatisfiability has been removed, so that trivially unsatisfiable formulae have been relegated out of the transition areas in our experiments.

---

11. The number of possible distinct modal atoms increases hyper-exponentially with $d$ (Horrocks et al., 2000).

**Artificiality:** Our method allows the user to shape the test formulae so that to maximize the resemblance to the expected typical inputs of his/her system(s). Of course, this is done within the limits imposed by randomness: the more irregular the typical input formulas, the higher the variance of the randomly generated formulas, the lower their average resemblance to the typical input formulas.

**Over-size:** The new method allows for generating extremely hard problems with reasonable size. It comes from the analysis of the resulting data that hard problems require very big amounts of both search branches and modal successors generated, so that the search is not dominated by parsing and data managing.

The generator presented by Horrocks and Patel-Schneider (2002), extends the 3CNF$_{\Box_m}$ generator of Giunchiglia et al. (2000) too. However, our new generator allows for shaping the probability distributions of both C and p, and for using different distributions at every depth level. In principle, the generator of Horrocks and Patel-Schneider (2002) allows also for setting the probabilities $n_p$ and $n_m$ by which propositional and modal atoms are negated. However, this feature is not used very much—in the experiments by Horrocks and Patel-Schneider (2002) $n_p$ is always 0.5 and $n_m$ is different from 0.5 only in one experiment— and adds nothing to the generality of the generator, so that in our new generator we decided not to re-introduce it.

### 7.3 Comparison with the QBF-based Method

Before comparing our new CNF$_{\Box_m}$ generation method with the QBF-based generation method, we must notice that, so far, they have been used in different ways, corresponding to the two different test techniques briefly summarized in Section 3.

- In the TANCS competition(s) (Massacci, 1999), the tests have been performed on single data points, and the results are presented in the form of big tables, each entry consisting of the number of successful solutions and in the rescaled geometrical mean CPU time for such solutions. Two or more systems are compared according to their number of successful solutions, considering the geometrical mean CPU time value only when the result is even. This is due to the fact that a comparison between geometrical means is possible only if they are computed on the same number of successful values, or, for a more accurate comparison, on the same successful values.[12] This method was chosen to guarantee the fairness of the comparison between the competitors, which is the key requirement in a competition.

- In this paper instead, we have focused on highlighting both the qualitative and quantitative behavior of the system(s). Thus we have preferred plots to tables, and we have preferred representing percentiles CPU times rather than the number of successful solutions and their geometrical mean times. In fact, the former does not require to distinguish between successful and non-successful solutions.[13] Thus, they are much more suitable for plotting, because a comparison on geometrical means makes sense only for those data points with the same number of successful solutions, which is very hard to follow in a plot.

---

12. In case of tests exceeding the timeout, geometrical means are altered by the truncation introduced by the unsuccessful solutions. Thus the geometrical mean makes sense only if calculated only on successful results.

13. If the percentage of successful solutions is greater or equal than $Q$, then the value $Q$-th percentile is not influenced by the truncation of values introduced by timeouts, otherwise it is equal to the timeout value.

Of course, both generators can be used in both ways. (See Heguiabehere and de Rijke (2001) for some plots with the random QBF-based method.) Comparing the two approaches above in organizing and presenting data is not one of the goals of this paper, so we restrict our analysis to the generation methods, independently from how they have been used so far.

The QBF-based generation method of Massacci (1999) shares with our new CNF$_{\Box_m}$ generation method several features —in particular **Scalability**, **Valid vs. not-valid balance**, **Termination**, **Reproducibility**, **Parameterization**, **Data Organization**, **Difficulty**, **Modal vs. propositional balance**, **Redundancy** and **Triviality**— for which considerations which are identical or analogous to those for our new method hold, once we consider parameters $V$, $D$ and $C$ instead of parameters $N$, $d$ and $L$. The following features instead deserve more discussion.

**Control:** The parameters $V$ and $D$ allow for controlling monotonically the difficulty of the test set. The parameter $C$ allows for controlling the satisfiability rate. However, unlike the CNF$_{\Box_m}$ case, the main parameters of the QBF generator (e.g., $D$ and $V$) do *not* have a direct meaning wrt. the main characteristics of the resulting modal formulae like, e.g., the modal depth and the number of propositional variables.

**Representativeness:** In general QBF formulae are good representatives for the whole class of quantified boolean formulae, as there is a way to convert a generic quantified boolean formula into QBF.[14] (The randomly generated QBF formulae used by Massacci (1999) restrict to those having a fixed amount of variables per alternation.) Nevertheless, the class of modal-encoded QBF formulae restrict to those having candidate Kripke structures with the very regular structure imposed by the QBF and/or binary search trees.

**Artificiality:** Unlike the CNF$_{\Box_m}$ case, the main parameters of the QBF generator (e.g., $D$ and $V$) do not have a direct meaning wrt. the main characteristics of the resulting modal formulae. Thus, it is hard to choose the parameters for the random QBF generator so that to resemble expected typical inputs of the system(s).

**Over-size:** One final problem with random modal-encoded QBF formulae is size. Initial versions of the translation method produced test sets in the 1GB range, which stressed too much the data-storage and retrieval portion of the provers. (For example, running DLP on these formulae resulted in a 1000s timeout without any significant search.) Although the encoding has been significantly improved in this sense, the current versions still produce very large modal formulae, mostly to constrain the Kripke structures.

Similar considerations have been very recently presented by Heguiabehere and de Rijke (2001).

On the whole, we believe that the QBF generation method is still appealing, and that the two methods can co-exist in any empirical test session.

---

14. Notice that by "QBF" here we denote the class of prenex CNF QBF formulae, given by an alternation of quantification variables ending with an existential one followed by a CNF propositional formula. The conversion works by lifting quantifiers outside the formula and then converting into k-CNF [k-DNF] the matrix if the innest quantifier is an ∃ [a ∀, negating the result and pushing down the negation recursively]. The conversion is truth-preserving [truth-inverting].

### 7.4 Complexity Issues

From a purely theoretical viewpoint, it is remarked that modal-encoded QBF formulae can capture the problems in $\Sigma_D^P$, while $\mathrm{CNF}_{\square_m}$ formulae are "stuck at NP" (Massacci, 1999)[15]. This statement requires some clarification.

First, test sets are necessarily *finite*, therefore it makes no sense to attribute to them a complexity class. Thus, when speaking of complexity classes for test problems, we do not refer to test sets, but rather to the *infinite* sets of formulae we could generate if we could have *unbounded* values for (at least one of) the generation parameters. In particular, the statement above means that the infinite set of QBF formulae with unbounded number of variables per alternation $V$ and bounded alternation depth $D$ is complete for $\Sigma_D^P$ (Garey & Johnson, 1979), while the infinite set of $\mathrm{CNF}_{\square_m}$ formulae with bounded depth and unbounded number of propositional variables is in NP (Halpern, 1995).

Secondly, the alternation depth $D$ and the variable number per alternation $V$ are not the "QBF-analogous" of $\mathbf{K}_{(\mathbf{m})}$'s modal depth and variable number respectively, as both the latter values for the resulting modal formulae grow as $O(D \cdot V)$.[16] In fact, QBF formulae with bounded alternation depth $D$ and unbounded number of variables per alternation $V$ give rise to modal formulae of both unbounded depth and unbounded number of variables.

Finally, the "$\Sigma_D^P$ vs. NP" issue of Massacci (1999) is not a matter of generators, but rather a matter of how such generators are used, and of how results are organized and presented. In fact, so far random $\mathrm{CNF}_{\square_m}$ testbeds have always been organized by fixing all the parameters except $L$ (modal depth $d$ included!) and making $L$ vary. This choice, whose goal is to produce data plots covering the satisfiability transition area, is what causes the testbed formulae to be "stuck at NP". To avoid this fact, one may want to make $d$ vary and to fix all the other parameters, as $\mathbf{K}_{(\mathbf{m})}$ satisfiability with unbounded depth and bounded number of propositional variables is PSPACE-complete (Halpern, 1995).

### 7.5 Asymptotic Behavior

Achlioptas et al. (1997) presented a study on the *asymptotic behavior* of random CSP problems. They showed that, for most well-known random generation models (which did not reveal flaws in empirical tests) the probability that problems are trivially unsatisfiable tends to 1 with $N \longmapsto \infty$, $N$ being the number of variables. Gent et al. (2001) lately explained this discrepancy between theoretical and empirical results by showing that the above phenomenon happens with significant probability only for values of $N$ which are out of the reach of current CSP solvers.

The problem is due to the possible presence of (implicit) *unary constraints* causing some variable's value to be inadmissible. If this occurs with some non-zero probability, then with non-zero probability some variable may have all its values inadmissible. This causes a "local" inconsistency of the whole problem, which is very easily revealed by the solver. When $N \longmapsto \infty$, the probability of not having such situation tends to zero. Analogous problems have been revealed with random SAT problems generated with the constant probability generation model, as unary clauses are gen-erated with non-zero probability (Mitchell et al., 1992), and with random QBF problems, as implicit unit clauses, —i.e., clauses containing only one existential variable— are generated with non-zero probability (Gent & Walsh, 1999). For the random k-SAT model, $k \geq 2$, such problem does not occur (Friedgut, 1998; Achlioptas et al., 1997).

---

15. More precisely, Massacci (1999) referred to the $3\mathrm{CNF}_{\square_m}$ formulae of Giunchiglia et al. (2000). The statement holds also for all the $\mathrm{CNF}_{\square_m}$ formulae.

16. As we have already noticed (Horrocks et al., 2000), a better "QBF-analogous" of the modal depth is the total number of universally quantified variables $U$ ($U = V \cdot \lfloor D/2 \rfloor$ in our case). In fact, like modal $\mathbf{K}_{(\mathbf{m})}$ with bounded depth, the class of QBF formulae with bounded $U$ is only complete in NP, as it is possible to "guess" a tree-like witness with $O(U \cdot 2^U)$ nodes.

Our generation model is far more complicated to analyze than the models above. First, $\text{CNF}_{\square_m}$ formulas have a much more complicated structure than random SAT, CSP and QBF formulas, involving a much wider number of parameters. Second, unlike with the models discussed above, the (constraints described by) $\text{CNF}_{\square_m}$ clauses are not picked in a *uniform* way, as the probability of generating a given $\text{CNF}_{\square_m}$ atom $\square_r \phi$ varies strongly with its depth and shape, and it is typically much smaller than that of generating a propositional atom $A_i$.[17] Thus, developing a formal probabilistic analysis for the asymptotic behavior of our model is out of the reach (and of the scope) of this paper. However, we provide here some heuristic considerations.

The simplest case is when we do not allow the generation of unary clauses at top level, that is, when `C = [[0,...],...]`, so that we do not have explicit unary constraints. We may still have *implicit* unary constraints like, e.g., $(A_i \vee \square_r \phi) \wedge (A_i \vee \neg\square_r \phi)$ or $(\square_i \psi \vee \square_r \phi) \wedge (\square_i \psi \vee \neg\square_r \phi)$. Anyway, a simple heuristic consideration suggests that, given the big numbers of distinct $\text{CNF}_{\square_m}$ modal atoms which may potentially be generated, such situations are more unlikely than that of having implicit unit constraints like $(A_i \vee A_j) \wedge (A_i \vee \neg A_j)$ in the standard 2-SAT model, which is free from the asymptotic local inconsistency problem.

A more critical case is when we allow for the generation of unary clauses at top level, that is, when `C = [[x,...],...], x > 0`. In this case we can generate unary clauses, and thus local inconsistencies, with non-zero probability. Thus, a simple way to avoid this problem is to restrict the values of $C$ so that not to allow unary top-level clauses, that is, to always set `C = [[0,...],...]`. Notice, however, that this hardly becomes a problem in practice if we respect the condition described in Sections 4.1 and 5.1 of avoiding purely propositional top-level clauses (that is, always set `p = [[...,0],...]`). In fact, given the big numbers of distinct $\text{CNF}_{\square_m}$ modal atoms which may potentially be generated, the probability of having two contradictory modal unit clauses $\square_r \phi$, $\neg\square_r \phi$ within the same formula becomes quickly negligible even with small depths.

Notice that here we have intentionally not considered "modal" implicit unary constraints like, e.g., $(A_i \vee \phi) \wedge (A_i \vee \psi)$, $\phi$ and $\psi$ being mutually inconsistent modal literals (e.g., $\phi = \square_r \varphi_1$, $\psi = \neg\square_r(\varphi_1 \vee \varphi_2)$). In fact, detecting such inconsistencies requires investigating recursively the modal successors, and therefore it is not "trivial".

## 8. Conclusions and Future Work

As shown by the test sets above our new method, in its basic form, allows us to generate a wider variety of problems covering more of the input space. We can better-tune the difficulty of problems for various parameter values, including the first reasonable test sets for maximum modal depths of 2 and 3. We can produce interesting scaling dimensions, varying more than just the number of propositional variables $N$. For example, we can now vary the propositional probability $p$ or the size of clauses $C$ to vary the difficulty of interesting problems. As neither $p$ nor $C$ are restricted to integral values, we have extremely fine control over the difficulty of test sets. Thus we can create more interesting test sets where the satisfiable/unsatisfiable transition is explorable with current decision procedures.

---

17. Again, we recall that the number of possible distinct $\text{CNF}_{\square_m}$ atoms increases hyper-exponentially with the modal depth $d$ (Horrocks et al., 2000).

We have drastically reduced the influence of trivial unsatisfiability, which flawed the previous $CNF_{\Box_m}$ methodologies when $p > 0$. We retain the desirable features of the previous $CNF_{\Box_m}$ methodologies. Our test sets are easy to reproduce and are not too large.

In our full methodology we have introduced the possibility of shaping the distribution of both the size and the propositional/modal rate of the clauses. This can be done at each level of modal depth. This allows for generating a much wider variety of problems, covering in principle the whole input space. For instance, we have produced a full test set with $d = 5$ and $N = 5$ (Figure 15).

We have not moved closer to application data, as there are no significant direct applications of modal decision procedures and thus no guidance for the sorts of inputs that would be close to application inputs. In any case, we believe we have moved closer than ever to the possibility of approximating given classes of input formulae.

There is still much work to be done using our generation methodology. We can produce more test sets and try these test sets out on various modal decision procedures. We may also want to uncover parameter settings where the full generality of our generation method is needed to produce reasonable test sets.

## Acknowledgments

We would like to thank Thomas Eiter and the three anonymous reviewers for their valuable comments and helpful suggestions which greatly improved the quality of the paper. The second author is supported by a MIUR COFIN02 project, code 2002097822_003, and by the CALCULEMUS! IHP-RTN EC project, contract code HPRN-CT-2000-00102, and has thus benefited of the financial contribution of the Commission through the IHP programme.

## Appendix A: Fully-detailed Proof of Theorem 6.1

**Theorem 6.1** *Let rnd_CNF$_{\Box_m}$ be a purely random generator as in Figure 13. Let $\varphi$ be a sorted CNF$_{\Box_m}$ formula of depth $d$ and with $L$ top-level clauses built on all the propositional atoms in $\{A_1, ... A_N\}$ and on all the modal boxes in $\{\Box_1, ... \Box_m\}$, which has no repeated clause at top level and no repeated atoms inside any clause at any level. Let $C$ and $p$ be built from $\varphi$ so that to verify points $(a)$ and $(b)$ of Section 6. Let $C'$ and $p'$ be obtained from $C$ and $p$ respectively by substituting some zero-values with some non-zero values. Then we have:*

(i) *rnd_CNF$_{\Box_m}$(d,m,L,N,p,C) returns $\varphi$ with some non-zero probability $\mathcal{P}$;*

(ii) *rnd_CNF$_{\Box_m}$(d,m,L,N,p',C') returns $\varphi$ with some non-zero probability $\mathcal{P}' \leq \mathcal{P}$.*

**Proof** The proof works by induction on the structure of $\varphi$. First, we prove that:

1. every propositional atom occurring in $\varphi$ at some depth $i$ is returned with the same non-zero probability $\mathcal{P}_1$ by both *rnd_atom(0,m,N,p,C)* and *rnd_atom(0,m,N,p',C')*;

2. every modal atom $\Box_v Cl$ occurring in $\varphi$ at some depth $i$ is returned with some non-zero probability $\mathcal{P}_2$ by *rnd_atom(d-i,m,N,p,C)*, and is returned with some non-zero probability $\mathcal{P}'_2 \leq \mathcal{P}_2$ by *rnd_atom(d-i,m,N,p',C')*;

3. every clause $Cl$ occurring in $\varphi$ at some depth $i$ is returned with some non-zero probability $\mathcal{P}_3$ by *rnd_clause(d-i,m,N,p,C)*, and is returned with some non-zero probability $\mathcal{P}'_3 \leq \mathcal{P}_3$ by *rnd_clause(d-i,m,N,p',C')*.

From point 3. we have that every top level clause $Cl_k$ is returned by *rnd_clause(d,m,N,p,C)* and *rnd_clause(d,m,N,p',C')* with some probabilities $\mathcal{P}_k$ and $\mathcal{P}'_k$ respectively, being $\mathcal{P}'_k \leq \mathcal{P}_k$. As $\varphi$ has no repeated clause, recalling a property of probabilities we have:

$$
\begin{aligned}
\mathcal{P} &= \begin{aligned}[t]
&\mathcal{P}_1 \cdot \\
&(\mathcal{P}_2 + \mathcal{P}_1\mathcal{P}_2 + \mathcal{P}_1^2\mathcal{P}_2 + \ldots) \cdot \\
&(\mathcal{P}_3 + (\mathcal{P}_1 + \mathcal{P}_2)\mathcal{P}_3 + (\mathcal{P}_1 + \mathcal{P}_2)^2\mathcal{P}_3 + \ldots) \cdot \\
&\ldots \\
&(\mathcal{P}_L + (\mathcal{P}_1 + \ldots + \mathcal{P}_{L-1})\mathcal{P}_L + (\mathcal{P}_1 + \ldots + \mathcal{P}_{L-1})^2\mathcal{P}_L + \ldots)
\end{aligned} \qquad (7)
\end{aligned}
$$

$$
= \prod_{k=1}^{L} \left( \sum_{i=0}^{\infty} (\sum_{s=1}^{k-1} \mathcal{P}_s)^i \cdot \mathcal{P}_k \right) \qquad (8)
$$

$$
= \prod_{k=1}^{L} \frac{\mathcal{P}_k}{1 - \sum_{s=1}^{k-1} \mathcal{P}_s}. \qquad (9)
$$

Notice that (8) is strictly monotonic in all its components. Thus, $\mathcal{P}' \leq \mathcal{P}$.

Now we need to prove points 1, 2 and 3.

1. Let $A_k$ be a propositional atom in $\{A_1, ..., A_N\}$ occurring in $\varphi$ at depth $i$, for some $i \leq d$. Then both *rnd_atom(0,m,N,p,C)* and *rnd_atom(0,m,N,p',C')* invoke *rnd_propositional_atom(N)*, which returns $A_k$ with probability $\mathcal{P}_1 = 1/N$.

2. Let $\Box_v Cl$ be a boxed clause occurring in $\varphi$ at depth $i$, for some $i < d$ and $v \leq m$. Then the clause $Cl$ occurs in $\varphi$ at depth $i+1$. (Notice that $i < d$ instead of $i \leq d$: $\Box_v Cl$ cannot occur in $\varphi$ at depth $d$, because $d$ is the maximum depth of $\varphi$.)

   (i) By inductive hypothesis, it follows from point 3. that $Cl$ is returned with some non-zero probability $\mathcal{P}_3$ by *rnd_clause(d-i-1,m,N,p,C)*. As $i < d$, *rnd_atom(d-i,m,N,p,C)* invokes *rand_box(m) · rand_clause(d-i-1,m,N,p,C)*, which returns $\Box_v Cl$ with the non-zero probability $\mathcal{P}_2 = 1/m \cdot \mathcal{P}_3$.

   (ii) By inductive hypothesis, it follows from point 3. that $Cl$ is returned with some non-zero probability $\mathcal{P}'_3 \leq \mathcal{P}_3$ by *rnd_clause(d-i-1,m,N,p',C')*. *rnd_atom(d-i,m,N,p',C')* invokes *rand_box(m) · rand_clause(d-i-1,m,N,p',C')*, which returns $\Box_v Cl$ with the non-zero probability $\mathcal{P}'_2 = 1/m \cdot \mathcal{P}'_3$. Thus, $\mathcal{P}'_2 \leq \mathcal{P}_2$.

3. Let $Cl$ be a clause with length $j$ and $r \leq j$ propositional literals, which occurs in $\varphi$ at depth $i$, for some $i \leq d$. As $\varphi$ is sorted, $Cl$ is represented as $Sort(\psi_1 \vee ... \vee \psi_r \vee \phi_1 \vee ... \vee \phi_{j-r})$, where $\psi_1, ..., \psi_r$ denote propositional literals and $\phi_1, ..., \phi_{j-r}$ denote modal literals.

   (i) By inductive hypothesis, it follows from point 1. that each propositional literal $\psi_k$ is returned with some non-zero probability $0.5 \cdot \mathcal{P}_{1,k}$ by *rnd_sign() · rnd_atom(0,m,N,p,C)*, and it follows from point 2. that each modal literals $\phi_l$ is returned with the non-zero probability $0.5 \cdot \mathcal{P}_{2,l}$ by *rnd_sign() · rnd_atom(d-i,m,N,p,C)*.

By construction of $C$, the $j$-th element of the $i$-th sublist in $C$ is non-zero; thus, $j$ is returned with some non-zero probability $\mathcal{P}_j$ by *rnd_length(d-i,C)*.

By construction of $p$, the $r+1$-the element of the $j$-th sub-sublist of the $i$-th sublist in $p$ is non-zero; thus, $r$ is returned with some non-zero probability $\mathcal{P}_{r|j}$ by *rnd_propnum(d,p,j)*.[18]

Similarly to (9), $\varphi$ has no repeated atoms inside any clause, so that $Cl$ is returned by *rnd_clause(d-i,m,N,p,C)* with the non-zero probability

$$\mathcal{P}_3 \; = \; \mathcal{P}_j \cdot \mathcal{P}_{r|j} \cdot (0.5)^j \cdot \prod_{k=1}^{r} \frac{\mathcal{P}_{1,k}}{1 - \sum_{s=1}^{k-1} \mathcal{P}_{1,s}} \cdot \prod_{l=1}^{j-r} \frac{\mathcal{P}_{2,l}}{1 - \sum_{t=1}^{l-1} \mathcal{P}_{2,t}}. \tag{10}$$

As with (9), the expression on the right in (10) is strictly monotonic in all its terms $\mathcal{P}_j$, $\mathcal{P}_{r|j}$, $\mathcal{P}_{1,k}$ 's, $\mathcal{P}_{2,l}$ 's within the domain of definition.

(ii) By inductive hypothesis, it follows from point 1. that each propositional literal $\psi_k$ is returned with some non-zero probability $0.5 \cdot \mathcal{P}'_{1,k} \leq 0.5 \cdot \mathcal{P}_{1,k}$ by *rnd_sign()* · *rnd_atom(0,m,N,p',C')*, and it follows from point 2. that each modal literals $\phi_l$ is returned with some non-zero probability $0.5 \cdot \mathcal{P}'_{2,l} \leq 0.5 \cdot \mathcal{P}_{2,l}$ by *rnd_sign()* · *rnd_atom(d-i,m,N,p',C')*.

By construction of $C$ and $C'$, the $j$-th element of the $i$-th sublist in $C'$ is non-zero; thus, $j$ is returned with some non-zero probability $\mathcal{P}'_j$ by *rnd_length(d-i,C')*. By construction of $C'$ from $C$, $\mathcal{P}'_j \leq \mathcal{P}_j$.

By construction of $p$ and $p'$, the $r + 1$-the element of the $j$-th sub-sublist of the $i$-th sublist in $p$ is non-zero; thus, $r$ is returned with some non-zero probability $\mathcal{P}'_{r|j}$ by *rnd_propnum(d,p,j)*. By construction of $p'$ from $p$, $\mathcal{P}'_{r|j} \leq \mathcal{P}_{r|j}$.

As $\varphi$ has no repeated atoms inside any clause, it follows that $Cl$ is returned by *rnd_clause(d-i,m,N,p',C')* with the non-zero probability

$$\mathcal{P}'_3 \; = \; \mathcal{P}'_j \cdot \mathcal{P}'_{r|j} \cdot (0.5)^j \cdot \prod_{k=1}^{r} \frac{\mathcal{P}'_{1,k}}{1 - \sum_{s=1}^{k-1} \mathcal{P}'_{1,s}} \cdot \prod_{l=1}^{j-r} \frac{\mathcal{P}'_{2,l}}{1 - \sum_{t=1}^{l-1} \mathcal{P}'_{2,t}}. \tag{11}$$

Because of the strict monotonicity of (10) and (11), we have that $\mathcal{P}'_3 \leq \mathcal{P}_3$.

Q.E.D.

18. Notice that $\mathcal{P}_{r|j}$ is a conditioned probability, that is, the probability of having $r$ propositional literal provided the clause has $j$ literals. This matches the fact that $j$ is an input in *rnd_propnum(d,p,j)*.